\RequirePackage{fix-cm}
\documentclass[smallextended]{svjour4}       % onecolumn (second format)
\smartqed  % flush right qed marks, e.g. at end of proof
\usepackage{graphicx}
\usepackage{array}
\usepackage{amsmath}
\usepackage{amsbsy}
\usepackage{amssymb}
\usepackage{algorithm}
\usepackage{algpseudocode}
\usepackage{float}
\usepackage{listings}
\usepackage{url}
\usepackage{hyperref}
\usepackage{enumitem}
\usepackage{adjustbox}

\usepackage{natbib}

\hypersetup{colorlinks,linkcolor={blue},citecolor={blue},urlcolor={blue}}

\floatstyle{ruled}
\newfloat{algorithm}{tbp}{loa}
\providecommand{\algorithmname}{Algorithm}
\floatname{algorithm}{\protect\algorithmname}

\usepackage{myfunctions_1.0}
\newcommand{\F}{\DATAMATRIX{F}{}}

\newcommand{\Ftilde}{\DATAMATRIX{\tilde{F}}{}}
\newcommand{\fij}{\DATAMATRIXelem{f}{i}{j}}
\newcommand{\ftildeij}{\DATAMATRIXelem{\tilde{f}}{i}{j}}

\newcommand{\subm}[3]{\DATAMATRIX{#1}{}_{\left[#2,#3\right]}}
\newcommand{\submZ}[2]{\subm{Z}{#1}{#2}}
\newcommand{\IntReprI}{\DATAMATRIX{H}{}}
\newcommand{\Mask}{\DATAMATRIX{V}{}}
\newcommand{\maskcj}{\DATAMATRIXelem{v}{c}{j}}
\newcommand{\GDstep}{\eta}
\newcommand{\lambdavector}{\LABELVECTOR{\lambda}{}}
\newcommand{\lambdavectorc}{\LABELVECTORelem{\lambda}{c}}
\newcommand{\partderiv}[2]{\frac{\partial #1}{\partial #2}}
\newcommand{\periodvector}{\mathbf{k}}

\usepackage[author={FMZennaro}]{pdfcomment}

\newcommand{\rev}[2]{#2}
\newcommand{\revII}[2]{#2}

\begin{document}

\title{On the Use of Sparse Filtering for Covariate Shift Adaptation%\thanks{Grants or other notes
}


\author{Fabio Massimo Zennaro         \and
        Ke Chen %etc.
}


\institute{
		F. M. Zennaro \at
           \email{zennarof@cs.manchester.ac.uk}           %  \\
           \and
           K. Chen \at
           \email{chen@cs.manchester.ac.uk}
           \and
           School of Computer Science \\
           The University of Manchester \\
           Kilburn Building\\
           Oxford Rd\\
           Manchester, M13 9PL\\ 
           United Kingdom           
}

\date{Received: date / Accepted: date}

\maketitle

\begin{abstract}
In this paper we formally analyse the use of sparse filtering algorithms to perform covariate shift adaptation. 
We provide a theoretical analysis of \emph{sparse filtering} by evaluating the conditions required to perform covariate shift adaptation. We prove that sparse filtering can perform adaptation only if the conditional distribution of the labels has a structure explained by a cosine metric. 
To overcome this limitation, we propose a new algorithm, named \emph{periodic sparse filtering}, and carry out the same theoretical analysis regarding covariate shift adaptation. We show that periodic sparse filtering can perform adaptation under the looser and more realistic requirement that the conditional distribution of the labels has a periodic structure, which may be satisfied, for instance, by user-dependent data sets. 
We experimentally validate our theoretical results on synthetic data. Moreover, we apply periodic sparse filtering to real-world data sets to demonstrate that this simple and computationally efficient algorithm is able to achieve competitive performances.

\end{abstract}

\section{Introduction \label{sec:Introduction}}

Traditional algorithms for machine learning often rely on the tacit
assumption of \emph{independent and identical distributed} (i.i.d.)
data. Under this assumption, the training samples on which an algorithm
is trained, and the test data on which the same algorithm is evaluated,
are taken to come from the same distribution. This assumption is critical for learning as it allows for a straightforward generalization of conclusions drawn from the training data to the test data. 
As long as the data can be modelled using this assumption, it is possible to provide statistical guarantees on the performance of traditional machine learning algorithms \citep{Vapnik1998,Bishop2006}. 
However, real-world
data rarely satisfy this assumption. Data collected from different
users, in different locations, or at different times do not fit the
requirement of coming from identical distributions. Not surprisingly, traditional machine learning algorithms proved to
perform poorly when trained and tested on data that are not identically
distributed.

Data that do not comply with the \iid assumption may be modelled according to the more realistic assumption of \emph{covariate shift}. The covariate shift assumption states that the marginal \rev{I put "feature" here among brackets, so it clarifies which distribution we are dealing with and it makes it clear that the term is optional and it will be dropped later.}{(feature)} distributions of training and test data are different, while the conditional distribution of the labels given the data is the same for both the training and test data \citep{Sugiyama2012}. \emph{Covariate shift adaptation} (CSA) aims at enabling supervised learning by compensating for the difference in the marginal distributions. \\

A common approach to CSA is through \emph{representation learning} (RL): given training and test data
exhibiting covariate shift, we try to learn a new representation of
the data that is not affected by covariate shift. 
A recently-proposed algorithm for RL is sparse filtering (SF) \citep{Ngiam2011}. SF learns maximally sparse representations of the data in an unsupervised way, aiming at learning a new useful distribution while overlooking the problem of explicitly modelling the noiseless marginal distribution that generated of the data. 
SF was shown to be an efficient and scalable algorithm, and it has been successfully applied to various real-world applications, leading to state-of-the-art performances even when paired with simple classifiers \citep{Ngiam2011}.\\ 
%SF has been used to deal with different forms of image recognition, such as iris recognition \citep{Raja2015} or vehicle recognition \citep{Dong2014}, or sound recognition \citep{Han2016a}.\\

So far, however, SF has never been used in the context of covariate shift. Here, we suggest that the disregard of SF for the problem of modelling the original marginal distribution of the data makes it a potential candidate for dealing with data affected by covariate shift. In addressing only the problem of shaping a useful distribution for the learned representations, SF may learn new representations in a way that is insensitive to covariate shift. SF may then offer an opportunity not only to avoid the computationally hard problem of learning the marginal distribution of the data, but also a way to circumvent the covariate shift problem. \\

In this paper we explore the potentialities and the limitations of using SF-like algorithms for CSA. We start by asking: \emph{can SF perform CSA?} In answering this question we will analyse the conditions required to perform CSA, and prove through a theoretical analysis that SF can indeed be used for CSA if the conditional distribution of the labels is explained by a cosine metric. Given this severe limitation, we move on to ask: \emph{can we devise a SF-based variant that can perform CSA under looser and more realistic conditions than SF?} We then develop an alternative algorithm based on SF, which retains the main advantages of SF while being able to perform CSA when the conditional distribution of the labels is periodic. We provide a theoretical analysis of this new algorithm and conduct experiments to demonstrate the usefulness of this SF-based algorithm on synthetic and real-world data.\\

The paper is organized as follows. 
Section \ref{sec:BG} formalizes the problem of learning under covariate shift and briefly reviews previous work on CSA and SF. 
Section \ref{sec:CSA-via-SF} presents a theoretical analysis of the conditions under which SF can perform CSA. 
Section \ref{sec:PSF} proposes a new SF-based algorithm to perform CSA. 
Section \ref{sec:CSA-via-PSF} provides a theoretical analysis of the conditions under which the new algorithm is guaranteed to perform CSA.
Section \ref{sec:Experiments} validates our theoretical results via an empirical study. Finally, Section \ref{sec:Conclusions} draws conclusions about our study.

\section{Background and Related Work \label{sec:BG}}

This section presents the concepts and the notation that will be used later in our theoretical study and in our experimental validations. Section \ref{sec:BG_Notation} introduces the formalism used to refer to the data and to the learning problem; Section \ref{sec:BG_CS} defines the problem of learning under covariate shift; Section \ref{sec:BG_CSA} discusses CSA and reviews related work; Section \ref{sec:BG_FDL} presents the SF algorithm and related work.

\revII{Corrected}{For clarity,} Table \ref{tab:Notation} summarizes the conventions and the notation adopted throughout this paper.

\begin{table*}
	\caption{Summary of notation. \label{tab:Notation}}
	\begin{centering}

\begin{tabular}{cl}
	\hline
	$a, b, c...$ & Generic scalars \tabularnewline
	$\APPLIEDFUNCTION{f}{\emptyarg}, \APPLIEDFUNCTION{g}{\emptyarg}, \APPLIEDFUNCTION{h}{\emptyarg}...$ & Generic functions \tabularnewline
	$\DATAVECTOR{a}{}, \DATAVECTOR{b}{}, \DATAVECTOR{c}{}...$ & Generic vectors \tabularnewline
	$\DATAVECTORelem{a}{i}$ & $i^{th}$ element of the the vector $a$ \tabularnewline
	$\DATAMATRIX{A}{}, \DATAMATRIX{B}{}, \DATAMATRIX{C}{}...$ & Generic matrices \tabularnewline
	$\DATAMATRIXrow{a}{i}$ & $i^{th}$ column of the the matrix $\DATAMATRIX{A}{}$ (sample) \tabularnewline
	$\DATAMATRIXcol{a}{j}$ & $j^{th}$ row of the the matrix $\DATAMATRIX{A}{}$ (feature) \tabularnewline
	$\DATAMATRIXelem{a}{i}{j}$ & $(i,j)^{th}$ element of the the matrix $\DATAMATRIX{A}{}$ \tabularnewline
%	$\domI, \domN, \domQ, \domR, \domC$ & Integer, natural, rational, real, complex domain \tabularnewline
%	$\domRgrez, \domRgrz$ & Non-negative, strictly positive domain \tabularnewline
	\hline
	$\absval{\emptyarg}$ & Absolute-value function \tabularnewline
	$\APPLIEDFUNCTION{\lO}{\emptyarg}, \APPLIEDFUNCTION{\lI}{\emptyarg}, \APPLIEDFUNCTION{\lII}{\emptyarg}, ..., \APPLIEDFUNCTION{\lp}{\emptyarg}$ & $\lO$-, $\lI$-, $\lII$-, $\lp$-norms \tabularnewline
	$\dist{A}{x}{y}$ & Distance $A$ between $x$ and $y$ \tabularnewline
	\hline \hline
	$\RV{X}{}$ & Random variable X \tabularnewline
	$\PDF{p}{X}$ & Probability distribution function. \tabularnewline
	 & Shorthand for $\PDF{p}{X=x}$ or $\PDF{p_X}{x}$. \tabularnewline
	$\PDF{p}{X,Y}$ & Joint distribution function \tabularnewline
	$\CONDPDF{p}{X}{Y}$ & Conditional distribution function \tabularnewline
	\hline
	$\mom{i}{X}$ & $i^{th}$ moment of the random variable $X$ \tabularnewline
	$\mean{X}$ & Expected value of the random variable $X$ \tabularnewline
	$\var{X}$ & Variance of the random variable $X$ \tabularnewline
	$\pdfdist{A}{p}{q}$ & $A$-distance between the pdfs $p$ and $q$ \tabularnewline
%	\hline
%	$\pdfgauss{\mu}{\mathbf{\Sigma}}$ & Gaussian pdf with mean $\mu$ and covariance $\mathbf{\Sigma}$ \tabularnewline
%	$\pdfunif{a}{b}$ & Uniform pdf within the interval $\closedinterval{a}{b}$ \tabularnewline
	\hline \hline
	$\nXsamples$ & Number of samples \tabularnewline
	$\nXfeatures, \nZfeatures$ & Number of features of original and learned representations \tabularnewline
	$\nYclasses$ & Number of labels \tabularnewline
	\hline
	$\X$ & Original data ($\X \in \Xspace$) \tabularnewline
	$\Z$ & Learned representations ($\Z \in \Zspace$) \tabularnewline
	$\Y$ & Labels ($\Y \in \Yspace$) \tabularnewline
	$\W$ & Weight matrix ($\W \in \Wspace$) \tabularnewline
	\hline
	$\domx$ & Domain of data \tabularnewline
	$\Xtr, \Xte, \Xta$ & Training, test, target data \tabularnewline
	$\rvX$ & Random variable for the original data \tabularnewline
	$\px$ & Pdf for the original data \tabularnewline
	$\rvXtr, \rvXte, \rvXta$ & Random variable for training, test,  target data \tabularnewline
	$\rvXfeatj$ & Random variable for the $j^{th}$ feature \tabularnewline
	\hline \hline
	
\end{tabular}
\end{centering}
\end{table*}

\subsection{Notation \label{sec:BG_Notation}}

Let us define the \emph{original data set} as the matrix $\X$ containing $\nXsamples$ data points. Each data point is represented as a vector $\x$ composed by $\nXfeatures$ features: $\x \in \domx \subseteq \origspace$. Elements of the vector $\x$ are referred as $\xij$, $1 \leq i \leq \nXsamples$ and $1 \leq j \leq \nXfeatures$. From a statistical point of view, a vector $\x$ is taken to be a random and independent sample from the multivariate marginal pdf $\px$.

Analogously, let us define the \emph{learned representation set} as the matrix $\Z$ containing the same number $\nXsamples$ of data points. Each data point is now represented as a vector $\z$ composed by $\nZfeatures$ features: $\z \in \domz \subseteq \learnspace$. Elements of the vector $\z$ are referred as $\zij$, $1 \leq i \leq \nXsamples$ and $1 \leq j \leq \nZfeatures$. From a statistical point of view, a vector $\z$ is taken to be a random and independent sample from the multivariate marginal pdf $\pz$.

Let us also consider the \emph{label set} as the vector $\Y$ made up of $\nXsamples$ categorical labels $\y \in \domy = \{1,2,\ldots,\nYclasses\}$. From a statistical point of view, a label $\y$ is taken to be a random and independent sample from the pdf $\py$.

In supervised learning, the original data is paired with the labels, $\set{\X,\Y}$, and learning means discovering a morphism $\MORPH{f}{\domx}{\domy}$ that maps original data onto the correct labels: $\APPLIEDMORPH{f}{\x}{\y}$. The degree of success of supervised learning is evaluated in terms of generalization, that is, how well the learned morphism $f$ approximates $\pygx$.
In unsupervised learning, learning may be expressed as the problem of modelling the original data through a morphism $\MORPH{g}{\domx}{\domz}$ that maps original data onto new representations: $\APPLIEDMORPH{g}{\x}{\z}$. The degree of success of unsupervised learning depends on the criteria chosen to evaluate the learned representations $\Z$ \citep{Bengio2012}. A simple way to assess the goodness of the learned representations is to evaluate their usefulness in supervised tasks, that is, evaluating whether approximating $\pygz$  is easier than approximating  $\pygx$.\\

For the sake of evaluating the generalization of our algorithms, we will follow the convention of partitioning our dataset in (at least) two subsets: a training data set $\Xtr$ made up of $\nXtrsamples$ samples from $\pxtr$ for learning the morphism $f$; and a test dataset $\Xte$ made up of $\nXtesamples$ samples from $\pxte$ for testing the morphism $f$.\\

When learning from data $\X$ in the space $\domx$, the results of many machine learning algorithm \revII{Corrected}{are} dependent on the metric assumed to explain the distances and the relationships between the samples \citep{Xing2003}.
\revII{I added this sentence to quote the source referenced by the reviewer. However I kept a quite general and informal tone because actually arguing when different metrics lead to the same result is not straightforward. While in general it is sound to assert that different metrics do NOT lead to the same result (as we stated in the previous sentence with reference to Xing), arguing when they actually lead to the same result is more complicated and it depends on the learning algorithm too. For many algorithms a simple topological equivalence may be enough; for other algorithms, who may be sensitive to the absolute value of the distances, a strong equivalence or an isomorphism may be required. Discussing these different cases seem to me outside our scope and it would require the introduction of a topological notation too. Therefore I tried to make a high-level statement with a reference to the source (Runde)}{Different choices of metrics may} lead to different learning outcomes unless the metrics may be proved to be equivalent \citep{Runde2007}.
Data $\X$ are often tacitly assumed to be defined in an Euclidean metric space, such that, given two points $\xone$ and $\xtwo$, we can compute their distance as $\dist{E}{\xone}{\xtwo}= \sqrt{\sumjfeatX \left( \xonej - \xtwoj \right)^2}$. The Euclidean distance $\dist{E}{\xone}{\xtwo}$ is then often assumed to define the structure of the data $\X$. However, different distance functions $\MORPH{D}{\domx \setCartprod \domx}{\domRgrez}$ may be considered to evaluate the distance between $\xone$ and $\xtwo$, thus inducing different structures on the data $\X$. 

When considering the conditional distribution $\pygx$, we say that the labelled data $\set{\X,\Y}$ \emph{has a structure explained by a metric function} $D$ if the following $ \epsilon$-$\delta$ condition is satisfied: given two data samples $\xone,\xtwo \in \domx$, if $\dist{}{\xone}{\xtwo} < \epsilon$,  then $\left| \CONDPDF{p}{y}{\xone} - \CONDPDF{p}{y}{\xtwo} \right| < \delta(\epsilon)$, where $\epsilon \in \domRgrz$ is an arbitrarily small positive constant, and $\MORPH{\delta}{\domRgrz}{\domRgrz}$ is a function returning an arbitrarily small positive constant depending on $\epsilon$.
In other words, a metric function $D$ explains the data structure if it preserves locally the conditional distribution: two points $\xone,\xtwo$ evaluated as close to each other, $\dist{}{\xone}{\xtwo} \approx 0$, have also similar conditional probabilities, $\CONDPDF{p}{y}{\xone} \approx \CONDPDF{p}{y}{\xtwo}$. \\
%Note that this assumption is consistent with the definition of the traditional smoothness assumption \citep{Bengio2012} under the distance is the Euclidean distance $\dist{E}{\xone}{\xtwo}$.

Lastly, given a data samples $\x$, we say that $\x$ is sparse if $m \ll \nXfeatures$ components of the vector $\x$ are active (that is, have a value different from zero), while the remaining $\nXfeatures-m$ components are inactive (that is, they have the value zero); we say that $\x$ is $k$-sparse if exactly $k$ components are active \citep{Makhzani2013}. 
By analogy, we may define sparsity for matrices (with reference to their components) and for random variables (with reference to their realizations).
With regard to matrices, given a data matrix $\X$ made up of $\nXsamples$ samples in $\nXfeatures$ dimensions, we also say that $\X$ is \emph{population spares} if the sample vectors $\x$ in $\X$ are sparse, while we say that $\X$ is \emph{lifetime sparse} or \emph{selective} if the feature vectors $\xj$ in $\X$ are sparse.

\subsection{Covariate Shift \label{sec:BG_CS}}

Under the \iid assumption all the data samples making up the matrix $\X$ are independent samples from the same pdf $\px$; thus, $\pxtr = \pxte$. As long as the conditional distribution of the labels remain the same across training and test data,  $\pygxtr = \pygxte$, then generalization from the training data $\Xtr$ to the test data $\Xte$ is straightforward \citep{Vapnik1998,Bishop2006}.

Unfortunately, the \iid assumption rarely holds in the real-world, where training data and test data may significantly differ in their distribution; this is the case, for instance, when an algorithm is trained on well-behaved data samples collected in a controlled environment, and then it is deployed in the wild. To model this situation we explicitly rely on the assumptions of \emph{covariate shift} \citep{Shimodaira2000,Sugiyama2012}. 
This assumption can be divided into two statements: (i) the marginal distribution of the training and test data are defined on the same sub-domains, $\domxtr = \domxte$, but their pdfs are different, $\pxtr \neq \pxte$, and (ii) the conditional distribution of the labels is the same for the training and test data, $\pygxtr = \pygxte$.
The first statement of the assumption of covariate shift is supposed to model the original problem of a difference between the distribution of training and test data \rev{I added this part to relate our problem of covariate shift to domain shift}{and to guarantee the identity of the sub-domains of the distributions; in case of domain shift, $\domxtr \neq \domxte$, it would not be possible to establish the identity of the conditionals because $\pygx$ would then be defined over different domains for training and test data.} 
The second statement is supposed to guarantee the possibility of learning and generalizing from training to test data; if both the marginal and the conditional distribution of the training and test data were to be different, generalization would not be possible as $\pxtr$ and $\pxte$ would be completely unrelated.\\ 

It is worth noticing that, even if many supervised discriminative algorithms are apparently concerned only with estimating a conditional distribution $\pygx$ which is the same over training and test data, they are actually affected by covariate shift in the marginals because they optimize an average error estimated on the training domain \citep{Yamada2012}. Indeed, standard supervised algorithms, such as SVM, are affected by the density of the training data; only purely conditional models, such as Gaussian processes, are immune to covariate shift \citep{Quinonero-Candela2009}. It is therefore important, even when learning a conditional distribution through standard supervised discriminative algorithms, to perform CSA to compensate for the difference between the marginal pdfs of the training and test data.

\subsection{Covariate Shift Adaptation \label{sec:BG_CSA}}

A CSA algorithm is an algorithm designed to make learning possible under covariate shift. Often, a CSA algorithm compensates for the difference between the training and test distributions in order to allow further supervised learning; in this case, a CSA algorithm should not only tackle covariate shift, but also retain relevant discriminative information carried by the conditional distributions.
We can express these requirements more formally in the following two necessary, but not sufficient, conditions:
\begin{description}
	\item[\emph{Marginal condition}:] a CSA algorithm must compensate for the difference between the marginal distributions of the training and the test data, that is, $\pdfdist{}{\pxtr}{\pxte} > \pdfdist{}{\pztr}{\pzte}$, where $\pdfdist{}{\emptyarg}{\emptyarg}$ is a measure of distance among 	pdfs, such as Kullback-Leibler divergence \citep{MacKay2003} or maximum mean discrepancy \citep{Gretton2012};
	\item[\emph{Conditional condition}:] a CSA algorithm must preserve the identity of the conditional distributions\footnote{This condition may actually be improved by requiring not just the preservation of the original $\pygx$ but instead the learning of a better-behaved conditional $\pygz$ (as it happens in representation learning).}, that is, $\pygz = \pygx$.
\end{description}
These two conditions have to be simultaneously satisfied to achieve good CSA. 
The first condition is necessary but not sufficient: if an algorithm were not to compensate for the difference in the distribution of training and test data, then no adaptation would take place; on the other hand, if an algorithm were to compensate for the difference in the distribution of training and test data, then we would still have no guarantee about CSA being successful since all discriminative information may be lost.
Similarly, the second condition, too, is necessary but not sufficient: if an algorithm were not to preserve the identity of the conditionals, then ensuing discrimination would be compromised; if an algorithm were to preserve the identity of the conditionals, then we would still have no guarantee that the distance between the marginals had been reduced.

Notice that these two conditions closely mirror the two statements in the assumption of covariate shift. The marginal condition requires the problem in the first statement of the covariate shift assumption to be addressed. The conditional condition requires that the guarantee for learning expressed in the second statement of the covariate shift assumption is preserved.\\

%We will consider two form of \emph{adaptation}: \emph{unsupervised CSA} carried out using the data $\{ \mathbf{X^{tr}}, \mathbf{X^{tar}} \}$, and  \emph{supervised CSA} carried out using the data $\{ \mathbf{X^{tr}}, \mathbf{Y^{tr}}, \mathbf{X^{tar}} \}$. 
%As further learning, we will consider only \emph{supervised classification} where learning is performed on $\{ \mathbf{X^{tr}}, \mathbf{Y^{tr}}\}$ and evaluation is performed on $\{ \mathbf{X^{tst}}\}$. Overall, all the CSA classification systems that we will consider will be trained on $\{ \mathbf{X^{tr}}, \mathbf{Y^{tr}}, \mathbf{X^{tar}} \}$ and evaluated on $\{ \mathbf{X^{tst}}\}$.

The statistical and machine learning literature contains several algorithms for performing CSA \citep[for a review of CSA algorithms see, for instance,][]{Jiang2008,Margolis2011}. Two types of approach to CSA are particularly relevant to our work: CSA via importance weighting and CSA via representation learning.

\subsubsection{CSA via Importance Weighting}
The first type of CSA is importance weighting (IW) \citep{Sugiyama2012}. IW rescales the loss function of a learning algorithm by the ratio between the distribution of the training data and the test data, thus emphasizing the contribution to learning of training points close to the test points and discounting the contribution of training points falling far from the test points. IW works under two conditions: (i) no domain shift, $\domxtr = \domxte$, which allows us to define the ratio between the pdf of the training samples and the pdf of the test samples; and (ii) smooth conditional distribution between the training and test domain \citep{Sugiyama2012}.
CSA is performed by rescaling the contribution to the loss function of each training sample according to the ratio between the marginal distribution of the training data and the test data. The loss function $\loss$ is modified based on each individual instance $\xtr$:
\[
\frac
{\PDF{p}{\rvXtr = \xtr}}
{\PDF{p}{\rvXte = \xtr}}
\APPLIEDFUNCTION{\loss}{\xtr}
\]
where the ratio $\frac{\pxtr}{\pxte}$ is estimated as $\frac
{\PDF {\estimation{p}} {\rvXtr} }
{\PDF {\estimation{p}} {\rvXte} }$ using density ratio estimation algorithms \citep{Sugiyama2012}.

The IW approach is theoretically well-grounded, but it has a high computational complexity due to the problem of estimating the pdfs $\pxtr$ and $\pxte$ or their ratio $\frac{\pxtr}{\pxte}$. The IW approach has been successfully applied to many problems, such as speaker identification, EEG processing, natural language processing \citep{Sugiyama2012} and emotional speech recognition \citep{Hassan2013}.

\subsubsection{CSA via Representation Learning}

The second type of CSA is representation learning (RL). In general, RL algorithms are designed to discover a new and better representation $\Z$ for the given data $\X$; better representations are intuitively described as abstract, denoised, robust or informative representations \citep{Bengio2012}. When applied to CSA, RL algorithms are enriched with the additional aim of tackling the shift in the pdfs of the training data and test data. This objective may be formally expressed in the requirement of discovering a new space $\domz$ such that $\pdfdist{}{\pxtr}{\pxte} > \pdfdist{}{\pztr}{\pzte}.$

Most of the existing RL algorithms work under the assumptions of \iid data and thus need to be redesigned to tackle covariate shift. However, some algorithms, most notably denoising auto-encoders (DAE) \citep{Vincent2008a}, have been shown to be able to perform effective CSA without any explicit modification to account for covariate shift.
A DAE module computes a new representation $\z$ of a data sample $\x$ as:
\[
\z = \APPLIEDFUNCTION {f} {\W \mathbf{\tilde{x}}_i+\mathbf{b}},
\]
where $\mathbf{\tilde{x}}_i$ is a corrupted noisy version of $\x$, $\MORPH{f}{\domR}{\domR}$ is an element-wise non-linear function, $\W \in \Wspace$ \revII{Corrected}{denotes} the generic weight matrix\footnote{Notice that the generic notation $\W$ is used to denote the weight matrix of different learning algorithms; the context disambiguates the meaning of the notation.} and $\mathbf{b} \in \learnspace$ is the bias vector. 
From the learned representation $\z$ a DAE module computes a reconstruction as:
\[
\mathbf{\hat{x}}_i = \APPLIEDFUNCTION {g} {\V \z + \mathbf{c}},
\]
where $\MORPH{g}{\domR}{\domR}$ is an element-wise non-linear function, $\V \in \Vspace$ is the reconstruction weight matrix and $\mathbf{c} \in \origspace$ is the reconstruction bias vector.
According to the implementation of the DAE, the non-linear functions might be the same, $\APPLIEDFUNCTION{f}{\emptyarg}=\APPLIEDFUNCTION{g}{\emptyarg}$, and the weight matrices might be tied, $\W = \transp{\V}$. 
Learning is performed by minimizing over all the parameters a reconstruction loss, such as the square error: 
\[
\argminproblem{\W\in\Wspace, \V\in \Vspace, \mathbf{b}\in\learnspace, \mathbf{c}\in\origspace}{\sumisamples \left( \mathbf{\hat{x}}_i - \x \right)^2}.
\]
A solution to this optimization problem can be computed by gradient descent minimizing over all the available data. Even if no rigorous theoretical explanation has been provided to justify the use of DAE for CSA, DAE and stacked DAE \citep{Vincent2010} were demonstrated to be able to perform effective CSA \citep{Glorot2011a}.\\

A representation learning solution explicitly designed to tackle covariate shift is sub-space alignment (SSA) \citep{Fernando2013}. SSA works under the assumption that the space defined by the PCA components of the training data can be projected  onto the space of the PCA components of the test data. SSA aligns the PCA spaces by computing the following representations:
\begin{eqnarray*}
	\Ztr & = & \Xtr \mathbf{T} \transp{\mathbf{T}} \mathbf{U}\\
	\Zte & = & \Xte \mathbf{U},
\end{eqnarray*}
where $\mathbf{T}$ is the matrix of the eigenvectors of the covariance matrix for the training data, and $\mathbf{U}$ is the matrix of the eigenvectors of the covariance matrix for the test data.\\

Other approaches to RL algorithms include methods based on the identification of pivot features \citep[such as \emph{structural correspondence learning},][]{Blitzer2006}, methods grounded in manifold learning \citep[such as \emph{geodesic flow kernels},][]{Gong2012}, and methods based on the minimization of a differentiable measure of distance between the original and the learned distribution \citep[such as \emph{kernel mean matching},][]{Huang2007, Quadrianto2009}.\\

CSA through RL allows to solve the problem of covariate shift, while, at the same time, learn a better-behaved conditional distribution $\pygz$. The RL approach has been successfully applied to many problems dealing with complex data, such as image recognition \citep{Kulis2011, Tzeng2014}, sentiment analysis \citep{Glorot2011a, Li2014} and emotional speech recognition \citep{Deng2014a}.

\subsection{Sparse Filtering \label{sec:BG_FDL}}
SF is the main representation learning algorithm belonging to the family of \emph{feature distribution learning} algorithms. While other unsupervised algorithms normally learn new representations by trying to recover the original noiseless distribution that generated the data, feature distribution learning algorithms aim at producing new representations
with a useful distribution for classification. SF learns
a mapping $ \MORPH{f}{\origspace}{\learnspace}$ projecting the data onto maximally sparse representation \citep{Ngiam2011}:
\[
\Z=\lIIcol { \lIIrow { \absval {\W \X} } },
\]
where $\W \in \Wspace$ is the weight matrix, 
$\absval{\emptyarg}$ is the element-wise absolute-value function, 
$\lIIrow{\emptyarg}$ is the $\lII$-normalization along the rows: $\lIIrow{\X} = \frac{\xij} {\sqrt {\sumisamples \left(\xij\right)^{2}}}$,
and 
$\lIIcol{\emptyarg}$ is the $\lII$-normalization along the columns: $\lIIcol{\X} = \frac{\xij} {\sqrt{ \sumjfeatZ \left(\xij\right)^{2}}}$.
For computational reasons, the absolute-value non-linearity is implemented as a soft absolute-value function: $\FUNCTION{f}{x}{\sqrt{x^2+\epsilon}}$, where $\epsilon$ is an arbitrarily small positive constant, such as $\epsilon=10^{-8}$. SF is trained by minimizing the  $\ell_1$-norm of the learned representations:
\[
\argminproblem{\W \in \Wspace} { \sumisamples \sumjfeatZ \zij},
\]
using gradient descent. 

SF was shown to be an excellent algorithm for unsupervised learning as it scales very well with the dimensionality of the input, it is easy to implement, it has essentially a single tunable parameter and it was able to achieve state-of-the-art performance on image recognition and phone classification \citep{Ngiam2011}. Since its introduction, SF has been integrated in several machine learning systems tackling problems as diverse as iris recognition \citep{Raja2015}, electrical fault detection \citep{Lin2016} or terrain classification \citep{Liu2016a}. 
Moreover, SF has also been the object of more theoretical studies \citep{Lederer2014, Zennaro2016}. In particular, in our earlier work \citep{Zennaro2016}, we showed that learning in SF can be explained in terms of defining filters in the original space $\origspace$ which map data samples $\x$ onto bases of the learned representation space $\learnspace$; consequently, we proved that SF can learn good representations for classification when the conditional distribution of the labels $\pygx$ is explained by a metric of cosine neighbourhoodness \citep{Zennaro2016}.

\section{Theoretical Analysis of Sparse Filtering for Covariate Shift Adaptation \label{sec:CSA-via-SF}}

This section provides a theoretical analysis of the standard SF algorithm aimed at understanding \emph{whether} and \emph{under which conditions} SF successfully performs CSA.
Section \ref{sec:CSA-via-SF_Marginal} discusses when the marginal condition for CSA is met by SF, while Section \ref{sec:CSA-via-SF_Conditional} discusses when the conditional condition for CSA is met.

\subsection{Marginal CSA Condition for SF \label{sec:CSA-via-SF_Marginal}}

In order to prove that SF meets the marginal condition for CSA, it is necessary to show that SF reduces the distance between the marginal distributions of the training and test data. Specifically, we need to show that the sub-domains of definition of the training and test data are kept the same, $\domztr = \domzte$, while the difference between the new pdfs $\pztr$ and $\pzte$ is reduced.\\ 

Let us start showing that, through its normalization steps, SF projects all the samples onto a common bounded sub-domain.

\begin{proposition} \label{prop:SF-Marginal-Condition-1}
	The sub-domain $\domz$ of the representations 	$\z$ learned by SF is $\closedinterval{0}{1}^{\nZfeatures}$.
\end{proposition}

\textbf{Proof sketch.} By definition, all the representations $\z$ learned by SF are bounded through $\ell_2$-normalization within the hyper-cube $\closedinterval{0}{1}^{\nZfeatures}$. For the complete proof, see Appendix \ref{app:Proof-of-Proposition-1}. $\QED$

Thus, independently from the original sub-domain of definition of the training and test data, the sub-domain $\domz$ of the learned representations of the training and test data is always identical.\\

Now let us consider the problem of a difference in the pdfs of training and test data. We show in the following proposition that SF addresses covariate shift by reshaping the pdfs of each learned feature within precise bounds.

\begin{proposition} \label{prop:SF-Marginal-Condition-2}
	For each learned feature $\zj$, the SF algorithm bounds $\mean{\rvZfeatj} \in \closedinterval{\epsilon}{1}$ and $\var{\rvZfeatj} \in \closedinterval{0}{1-\epsilon^2}$, where $\epsilon>0$ is an arbitrarily small value defined in the non-linearity of SF. 
	Moreover, if we make the assumption that learned representations are $\closedinterval{1}{k}$-sparse in population and lifetime, and that $\epsilon$ is negligible, then we have the bounds $\mean{\rvZfeatj} \in \closedinterval{\frac{1}{\nXsamples}}{\frac{k}{\nXsamples}}$ and $\var{\rvZfeatj} \in \closedinterval{\frac{\nXsamples-k^{2}}{\nXsamples^{2}}}{\frac{\nXsamples k-1}{\nXsamples^{2}}}$.
\end{proposition}

\textbf{Proof sketch.} By closely analysing each processing step of SF, we can bound the distribution of each feature. For the formal analysis, see Appendix \ref{app:Proof-of-Proposition-2}. $\QED$

Thus, SF tackles the problem of covariate shift by forcing all the features to have bounded expected values and bounded variances. These bounds on the distribution of the features $\rvZfeatj$ are theoretically independent of the data matrix being processed. The fact that the new learned distributions $\PDF{p}{\rvZtr_{\emptyarg,j}}$ and $\PDF{p}{\rvZte_{\emptyarg,j}}$ have the first statistical moments similarly bounded on the same interval suggests that SF is able, at least in part, to mitigate the problem of covariate shift.

More interestingly, Proposition \ref{prop:SF-Marginal-Condition-2} reveals that, under the assumption of $k$-sparsity, SF moves the center of mass of the pdf of each feature $\PDF{p}{\rvZfeatj}$ towards zero, and it also decreases the variance in proportion to the number of samples $\nXsamples$. In other words, SF not only shapes the overall pdf $\pz$ towards being mainly localized around zero, but also does the same for the individual pdf of each feature $\PDF{p}{\rvZfeatj}$.  

Note that this is consistent with the interpretation of SF in terms of entropy minimization \citep{Zennaro2016}; the maximization of sparsity may be interpreted as a proxy for the minimization of entropy \citep{Principe2010,Pastor2015}, and SF can be understood as an algorithm projecting the original data onto representations with a pdf with minimal entropy. Therefore, independently from the original pdfs $\pxtr$ and $\pxte$, SF aims at learning a new representation with an entropy-minimized pdf $\pz$.

\subsection{Conditional CSA Condition for SF \label{sec:CSA-via-SF_Conditional}}

Satisfying the marginal condition for CSA is not enough to guarantee that SF always generates useful representations for classification. Indeed, if the samples were to be randomly mapped to maximally sparse representations in agreement with the marginal condition only, we would very likely lose all discriminative information carried by the data. For discriminative information to be retained, SF must satisfy the conditional condition requiring the preservation of the identity of the conditional distribution $\pygz = \pygx$.
To verify when this condition is met, we need to determine in which situation SF can preserve the identity of the conditional distributions of the labels given training and test data.

In our earlier work \citep{Zennaro2016}, we showed that SF can preserve the conditional structure of $\pygx$ explained by a metric of cosine neighbourhoodness (\revII{Corrected}{in order to be self-contained, we include} in Appendix \ref{app:preservation-cosine-neigh} the proof of the theorem proving the preservation of cosine neighbourhoodness in SF appearing in our previous work, \citet{Zennaro2016}). By expressing cosine neighbourhoodness in terms of cosine distance, it is immediate to derive the following corollary for the preservation of the structure of the conditional distribution $\pygx$.

\begin{corollary} \label{coroll:SF-Conditional-Condition}
	SF preserves the structure of the conditional distribution $\pygx$ explained by the metric of cosine distance $\dist{C}{\xone}{\xtwo} = 1 - \frac{\xone \xtwo}{\APPLIEDFUNCTION{\lII}{\xone}\APPLIEDFUNCTION{\lII}{\xtwo}}$.
\end{corollary}

SF transforms the conditional distribution $\pygx$ explained by the cosine metric in the original space into a new conditional distribution $\pygz$ explained by the Euclidean distance in the learned space \citep{Zennaro2016}. Thus, if in the original space a small cosine distance implies a small difference in the conditional distribution:
\[
\left[\dist{C}{\xone}{\xtwo} < \epsilon \right] \implies
\left[\absval{\CONDPDF{p}{y}{\xone} - \CONDPDF{p}{y}{\xtwo}} < \APPLIEDFUNCTION{\delta}{\epsilon}\right],
\]
where $\epsilon$ is an arbitrarily small value $\epsilon>0$ and $\APPLIEDFUNCTION{\delta}{\epsilon}$ is a function dependent from $\epsilon$ and returning an arbitrarily small value $\APPLIEDFUNCTION{\delta}{\epsilon}>0$, then in the learned space a small Euclidean distance implies a small difference in the conditional distribution:
\[
\left[\dist{E}{\zone}{\ztwo} < \epsilon' \right] \implies
\left[\absval{\CONDPDF{p}{y}{\zone} - \CONDPDF{p}{y}{\ztwo}} < \APPLIEDFUNCTION{\delta'}{\epsilon'}\right].
\]

This, in turn, allows standard Euclidean-based classifiers to successfully process the new representations.\\

In conclusion, bringing together these results for CSA using SF, we have:

\begin{theorem}
	SF meets the necessary and sufficient conditions for CSA if the structure of the conditional distribution $\pygx$ is explained by a cosine metric.
\end{theorem}

\textbf{Proof.} This theorem follows directly from Propositions \ref{prop:SF-Marginal-Condition-1} and \ref{prop:SF-Marginal-Condition-2}, proving the marginal condition, and Corollary \ref{coroll:SF-Conditional-Condition}, proving the conditional condition. $\QED$\\

SF is then able to perform some degree of CSA under the same conditions required for SF to perform useful unsupervised learning \citep{Zennaro2016}. SF can then be used to perform CSA, but only on a limited number of real-world data sets that can be expected to comply with the assumption of a conditional structure explained by the cosine metric.

\section{Periodic Sparse Filtering \label{sec:PSF}}

In the previous section we showed that the capacity of SF to perform CSA is limited by a strict requirement on the conditional distribution. In this section we propose a new SF-based algorithm that extends this requirement to a conditional distribution explained by a generic periodic structure. A periodic conditional structure could be used to model several real-world scenarios, as periodic functions would allow us to capture common regularities present in the marginal distributions of training and test data. For instance, in the common case of user-dependent data, we could model each user as described by a specific pdf $\PDF{p}{X^{user_i}}$ on a restricted sub-domain $\domx^{user_i}$; labels may then be expected to show some degree of regularity over each sub-domain, such that $ \CONDPDF{p}{Y}{X^{user_1}} = \CONDPDF{p}{Y}{X^{user_2}} = \dots = \CONDPDF{p}{Y}{X^{user_i}}$; learning the periodic behaviour over the set of training users would then allow us to generalize it to the set of test users.\\

We set out to define a new SF-based algorithm, which we call \emph{periodic sparse filtering} (PSF), by enriching the original SF algorithm with two crucial properties: (i) the ability to generate periodic filters in the original space, and (ii) the ability to capture the periodic conditional structure underlying the data from available labels.

The first property is required in order to preserve the periodic structure of the conditional distribution $\pygx$.
SF generates hyper-conical filters that can capture a radial structure, but are unable to model more complex periodic structures in the original space $\origspace$ \citep[see Section 3.11 in][]{Zennaro2016}. In order to capture periodic structures, we substitute the absolute-value non-linearity with a sinusoidal function which can generate periodic filters that regularly tile the whole original space. 
The new transformation in PSF is defined as follows:
\[
\Z= \lIIcol{\lIIrow{\APPLIEDFUNCTION{g}{\W \X}}},
\]
where $\APPLIEDFUNCTION{g}{x}$ is a positive element-wise sinusoidal function, such as $1+\epsilon+\sin(x)$ or $1+\epsilon+\cos(x)$, a sinusoidal function shifted by $1+\epsilon$, with $\epsilon>0$ being an arbitrarily small constant, such as $\epsilon=10^{-8}$, that guarantees the strict positivity of the output of $\APPLIEDFUNCTION{g}{x}$.

The second property is required in order to capture the correct periodicity underlying the conditional distribution $p\left(Y\vert X\right)$.
Given a set of unlabelled data, it is possible to discover different periodic functions underlying the data; however, only few of these periodic functions can be usefully related to the conditional distribution $\pygx$. In order to direct the algorithm to discover the conditional periodic function of interest, we turned the original unsupervised adaptation algorithm into a supervised adaptation algorithm. 
Let $\Xtr$ be the training data with its associated labels $\Ytr$. Assuming that we are learning a new representation in $\learnspace$, let us partition the $\nZfeatures$ learned features in $\nYclasses+1$ groups with arbitrary cardinality, corresponding to the $\nYclasses$ classes defined in $\Ytr$, plus one group for potentially unlabeled samples.
We can then re-define the learned representation matrix $\Z$ using the following block matrix notation:
\[
\Z=\left[\begin{array}{ccccc}
\submZ{1}{1} & 	\submZ{1}{2} & \dots & \submZ{1}{C} & \submZ{1}{C+1}\\
\submZ{2}{1} & 	\submZ{2}{2} & \dots & \submZ{2}{C} & \submZ{2}{C+1}\\
\dots & \dots & \dots & \dots & \dots\\
\submZ{C}{1} & 	\submZ{C}{2} & \dots & \submZ{C}{C} & \submZ{C}{C+1}\\
\submZ{C+1}{1} & \submZ{C+1}{2} & \dots & \submZ{C+1}{C} & \submZ{C+1}{C+1}\\
\end{array}\right],
\]
where $\submZ{i}{j}$ is the block matrix containing the $i^{th}$ group of learned features from the samples belonging to the $j^{th}$ class. This structure highlights the contribution of each group of learned features to the representation of samples belonging to a given class.
Exploiting the representation matrix in this new form, we can redefine the loss function of PSF as:
\[
\argminproblem{\W \in \Wspace}{\APPLIEDFUNCTION{\lI}{\Z} - 
	\sum_{c=1}^{\nYclasses} } \lambda_{c} \multiplication \APPLIEDFUNCTION{\lI}{\submZ{c}{c}},
\]
where $\lambda_{c} \in \domR$ is a scaling factor. The first term of this new loss function is the same as in SF, and its aim is to push for learning sparse representations. The second term of the loss function is the PSF addition. This term has an opposite effect compared to the first: while the first term tries to reduce and shrink the values of the elements of $\Z$, the second term tries to increase the values of the sub-matrices of $\Z$ around the diagonal. Overall, this loss function should push away mass from the components off the diagonal and push it onto the components on the diagonal. These dynamics are reminiscent of learning with energy-based models \citep{LeCun2006}, in which an energy surface over a learned space is pulled down over desired outcomes and pulled up over other outcomes; similarly, even though in a reverse way, our loss function tries to increase the mass over certain outcomes and decrease it over other outcomes. Practically, the learning algorithm is now biased towards generating sparse representations where the $c^{th}$ group of learned features tends to activate for the samples belonging to the $c^{th}$ class.\\

The pseudo-code for the new PSF algorithm\footnote{The Python source code is available online at: \url{https://github.com/FMZennaro/PSF}} is provided in Algorithm \ref{alg:PSF}. In conclusion, our PSF algorithm retains the simplicity of the original SF, but at the same time it exploits the labels available for classification to perform CSA. 

\begin{algorithm} \caption{Periodic Sparse Filtering (PSF) \label{alg:PSF}} 
	\begin{algorithmic}[1] 
		\Statex \textbf{Input:} training data $\Xtr$; training labels $\Ytr$; target data for adaptation $\Xta$.
		\Statex \textbf{Hyper-parameters:} learned dimensionality $\nZfeatures$; lambda vector $\lambdavector$; binary matrix $\Mask$ defining the block matrix structure of $\Z$ such that $\maskcj=1$ if the $j^{th}$ learned feature is activated by the $c^{th}$ class; gradient descent step $\GDstep$.
		\Statex 
		\State $\X \leftarrow \Xtr \setunion \Xta$ 
		\State $\W \leftarrow \textnormal{initialize each weight as } \pdfstdnorm$
		\State $C \leftarrow \#$classes in $\Ytr$ \Statex 
		\Repeat  	
		\State $\IntReprI \leftarrow \W \X$ 	
		\State $\F \leftarrow 1 + \epsilon + \sin\IntReprI$ 
		\State $\Ftilde \leftarrow \frac{\fij}{\sqrt{\sumisamples\fij^2}}$ 	
		\State $\Z \leftarrow \frac{\ftildeij}{\sqrt{\sumjfeatZ\ftildeij^2}}$ 	
		\State $\loss_1 \leftarrow \sumisamples \sumjfeatZ \zij$
		\State $\loss_2 \leftarrow \sum_{c=1}^{\nYclasses} \sum_{i: \y=c} \sum_{j: \maskcj=1 } \lambdavectorc \multiplication \zij$
		\State $\loss \leftarrow \loss_1 - \loss_2$ 	
		\Statex 	
		\State $\W \leftarrow \W - \GDstep \nabla \loss$ 
		\Until termination condition for gradient descent is met 
		\Statex 
		\Return $\Z$ 
	\end{algorithmic} 
\end{algorithm}

\section{Theoretical Analysis of Periodic Sparse Filtering for Covariate Shift Adaptation \label{sec:CSA-via-PSF}}

This section offers a theoretical analysis of PSF, analogous to the one we conducted on SF in Section \ref{sec:CSA-via-SF}.
Section \ref{sec:CSA-via-PSF_Marginal} discusses when the marginal condition for CSA is met by PSF, while Section \ref{sec:CSA-via-PSF_Conditional} discusses when the conditional condition is met.

\subsection{Marginal CSA Condition for PSF \label{sec:CSA-via-PSF_Marginal}}

Analogously to the theoretical analysis of the marginal CSA condition for SF, we first prove that PSF projects all the samples onto a common bounded sub-domain. It is straightforward to show that PSF guarantees this property in the same way as SF does.

\begin{proposition} \label{prop:PSF-Marginal-Condition-1}
	The sub-domain $\domz$ of the representations $\z$ learned by PSF is $\closedinterval{0}{1}^{\nZfeatures}$.
\end{proposition} 

\textbf{Proof.} This proposition follows from the $\lII$-normalization along the rows, which is exactly the same in both SF and PSF.
Therefore, the proof for Proposition \ref{prop:SF-Marginal-Condition-1} holds here as well. $\QED$\\

Next, let us consider the problem of a difference in the pdfs of training and test data. Like SF, PSF also reshapes the pdfs of each learned feature within precise bounds as specified in the following proposition.

\begin{proposition} \label{prop:PSF-Marginal-Condition-2}
	For each learned feature $\zj$, the PSF algorithm bounds $\mean{\rvZfeatj} \in \closedinterval{\epsilon}{1}$ and $\var{\rvZfeatj} \in \closedinterval{0}{1-\epsilon^2}$, where $\epsilon>0$ is an arbitrarily small value defined in the non-linearity of PSF. 
	Moreover, if we make the assumption that learned representations are $\closedinterval{1}{k}$-sparse in population and lifetime, and that $\epsilon$ is negligible, then we have the bounds $\mean{\rvZfeatj} \in \closedinterval{\frac{1}{\nXsamples}}{\frac{k}{\nXsamples}}$ and $\var{\rvZfeatj} \in \closedinterval{\frac{\nXsamples-k^{2}}{\nXsamples^{2}}}{\frac{\nXsamples k-1}{\nXsamples^{2}}}$.
\end{proposition}

\textbf{Proof.} The bounds on the distribution of the learned features depends on the $\lII$-normalization steps, which are the same in SF and PSF.
Therefore the proof for Proposition \ref{prop:SF-Marginal-Condition-2} holds here as well. $\QED$

\subsection{Conditional CSA Condition for PSF \label{sec:CSA-via-PSF_Conditional}}

Having proved that PSF meets the marginal condition to perform CSA, we now show that our PSF further meets the conditional condition for CSA when data has a periodic structure.

By construction, the PSF algorithm was designed with the idea of preserving a periodic structure underlying the data. It is then natural to expect that the identity of the conditional distributions, $\pygz=\pygx$, is preserved when the data exhibit such a structure. To confirm that our design works as we intended, we first prove a theorem about data structure preservation in PSF. 

\begin{theorem} 
	Let $\xone \in \origspace$ be a point in the original space and let $\zone \in \learnspace$ be its corresponding representation learned by PSF. Then there is an infinite set of points $\x \in \origspace$ that map onto the same representation $\zone$. The set of the points $\x \in \origspace$ built from $\xone$ with period $\W{}^{-1} \periodvector \pi$, where $\W$ is the weight matrix of PSF and $\periodvector$ is a vector of integer constants in $\domz$, is included in this set.
\end{theorem}

\textbf{Proof sketch.} By analysing each processing step of PSF, we can reconstruct how periodic filters are defined in the original space and show that points falling within these filters are mapped onto identical representations. For the complete proof, see Appendix \ref{app:Proof-of-Theorem-2}. $\QED$\\

\begin{figure}
	\begin{centering}
		\includegraphics[scale=0.4]{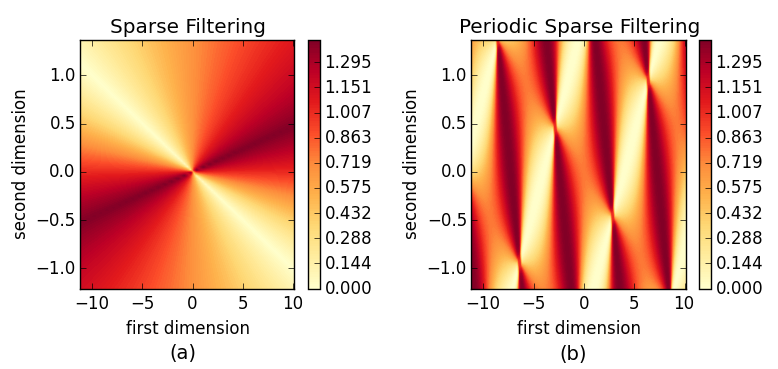}
		\par\end{centering}
	\caption{Sample filters in the original space $\domR^2$ learned by: (a) SF and (b) PSF \label{fig:filters}}
\end{figure}

This theorem proves that the PSF algorithm can define a specific frequency for each dimension in the original space; all the points with coordinates that are multiples of these frequencies are then mapped onto the same representation (see, for illustration, a comparison between the filters learned by SF and PSF in Figure \ref{fig:filters}(b)).\\

From Theorem 2, we can immediately derive the following corollary on the preservation of the structure of the conditional distribution $\pygx$.
\begin{corollary} \label{coroll:PSF-Conditional-Condition}
	PSF preserves the structure of the conditional distribution $\pygx$ explained by a periodic metric  $\dist{P}{\xone}{\xtwo} = \APPLIEDFUNCTION{\lp}{\APPLIEDFUNCTION{g_\periodvector}{\xone} - \APPLIEDFUNCTION{g_\periodvector}{\xtwo}}$, where $\APPLIEDFUNCTION{g_\periodvector}{\x}$ is an element-wise periodic function with periods $\periodvector$ and $\APPLIEDFUNCTION{\lp}{\emptyarg}$ is an $\lp$-norm.
\end{corollary}

In this case, if in the original space a small periodic distance implies a small difference in the conditional distribution:
\[
\left[\dist{P}{\xone}{\xtwo} < \epsilon \right] \implies
\left[\absval{\CONDPDF{p}{y}{\xone} - \CONDPDF{p}{y}{\xtwo}} < \APPLIEDFUNCTION{\delta}{\epsilon}\right],
\]
then, in the learned space a small Euclidean distance implies a small difference in the conditional distribution:
\[
\left[\dist{E}{\zone}{\ztwo} < \epsilon' \right] \implies
\left[\absval{\CONDPDF{p}{y}{\zone} - \CONDPDF{p}{y}{\ztwo}} < \APPLIEDFUNCTION{\delta'}{\epsilon'}\right].
\]
\\
In conclusion, we can summarize our theoretical analysis of PSF for CSA in the following theorem:

\begin{theorem} \label{thm:preservation-periodicity}
	PSF meets the necessary and sufficient conditions for CSA if the structure of the conditional distribution $\pygx$ is explained by a periodic metric.
\end{theorem}

\textbf{Proof.} This theorem follows directly from Propositions \ref{prop:PSF-Marginal-Condition-1} and \ref{prop:PSF-Marginal-Condition-2}, proving the marginal condition, and Theorem \ref{thm:preservation-periodicity} and Corollary \ref{coroll:PSF-Conditional-Condition}, proving the conditional condition. $\QED$\\

Compared to SF, PSF can then perform CSA under the looser requirement of a periodic structure. However this increased capacity requires additional information for directing the learning process; since many different periodic structures may be learned, PSF needs to rely on side information in the form of labels in order to extract a useful or meaningful structure.

\section{Experimental Validation \label{sec:Experiments}}

In this section we validate and test our theoretical results on the use of SF and PSF for CSA. 
In Section \ref{sec:Synthetic} we start by running experiments on synthetic data sets in order to obtain a better understanding and to be able to easily visualize the effects of covariate shift and the contributions of SF and PSF. In Section \ref{sec:Real-World-Data-Experiments} we execute a series of experiments on real data sets in which we measure the effectiveness of SF and PSF against other CSA algorithms used in the machine learning literature.\\

In order to highlight the dependency of CSA algorithms on their underlying assumptions about the structure of the data and to measure SF and PSF against state-of-the-art methods, we have undertaken a comparative study using the three CSA algorithms reviewed in Section \ref{sec:BG_CSA}. We implemented the following classification systems, using a linear support vector machine (SVM) as a default classifier, unless otherwise stated:

\begin{enumerate}[label=(\roman*)]
	
	\item \emph{SVM (without CSA)}: this model does not perform any CSA and it is used only to provide a baseline against which to evaluate the contribution of CSA algorithms.
	
	\item \emph{SF+SVM}: this model allows for the evaluation of the CSA capacity of SF; as discussed in Section \ref{sec:CSA-via-SF}, SF is supposed to successfully perform CSA when the data has a radial structure.
	
	\item \emph{PSF+SVM}: this model allows for the evaluation of the CSA capacity of PSF; as discussed in Section \ref{sec:CSA-via-PSF}, PSF is supposed to successfully perform CSA when the data has a periodic structure.
	
	\item \emph{IW+LSPC}: this model implements a commonly used setting for CSA based on IW \citep{Hachiya2012} and relies on a least-square probabilistic classifier (LSPC) \citep{Sugiyama2012a}; as recalled in Section \ref{sec:BG_CSA}, IW can be expected to successfully perform CSA in absence of domain shift, $\domxtr = \domxte$, and when the conditional distribution $\pygx$ is smooth \citep{Sugiyama2012a}. 
		
	\item \emph{SSA+SVM}: this model implements the SSA algorithm \citep{Fernando2013}; as explained in Section \ref{sec:BG_CSA}, SSA can be expected to successfully perform CSA when the PCA components of the training and the test data can be projected on each other \citep{Fernando2013}. 
	
	\item \emph{DAE+SVM}: this model implements the DAE algorithm, using a setting analogous to the one considered, for instance, in \citep{Deng2014c}; while DAE has been successfully used for CSA as discussed in Section \ref{sec:BG_CSA}, the conditions under which DAE successfully perform CSA have not yet been derived theoretically. 
	
\end{enumerate}
More details on the implementation and configuration of the above models are provided in Appendix \ref{app:Experiments}.

\subsection{Synthetic Data Set \label{sec:Synthetic}}

\begin{figure*}
	\begin{centering}
		
		\includegraphics[scale=0.45]{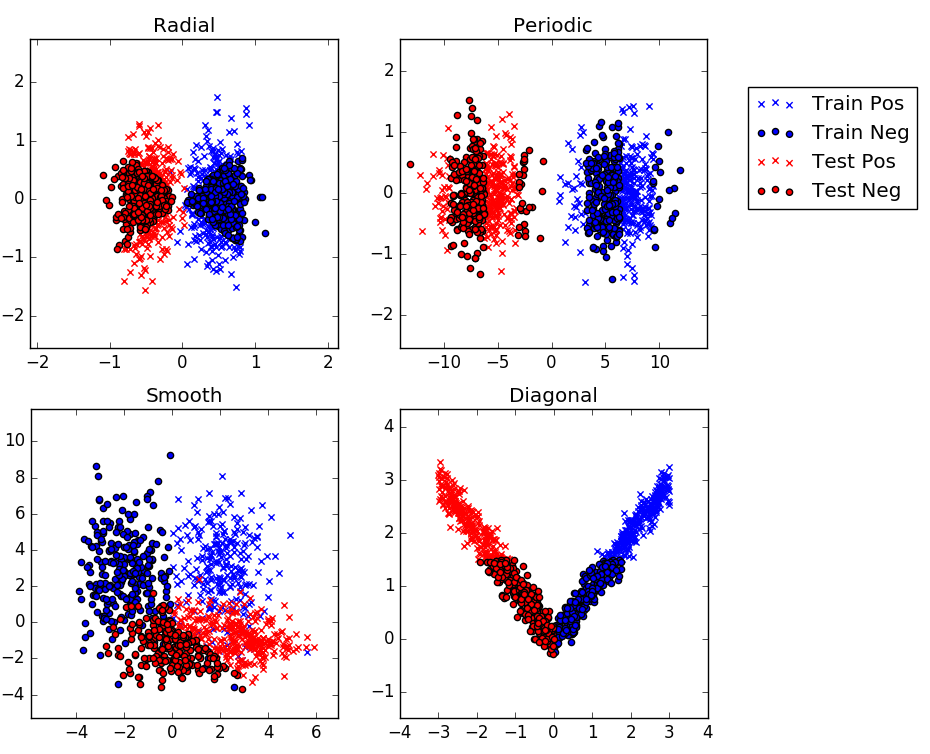}
		\caption{Synthetic data sets \emph{radial}, \emph{periodic},  \emph{smooth} and \emph{diagonal}, built respectively on the CSA assumptions of SF, PSF, IW and SSA. \label{fig:data}}
		%\emph{Radial}: $p\left( X^{(tr)}\right)$ and $p\left( X^{(tr)}\right)$ are two different Gaussian distributions; $p \left( X \vert Y \right)$ is explained by a deterministic radial function. \emph{Periodic}: $p\left( X^{(tr)}\right)$ and $p\left( X^{(tr)}\right)$ are two different Gaussian distributions; $p \left( X \vert Y \right)$ is explained by a deterministic periodic function. \emph{Smooth}: $p\left( X^{(tr)}\right)$ and $p\left( X^{(tr)}\right)$ are two mixtures of Gaussian distributions; $p \left( X \vert Y \right)$ is explained by a deterministic function; this data set is the same data set used in \citep{Hachiya2012} to show the effectiveness of IW. \emph{Diagonal}: $p\left( X^{(tr)}\right)$ and $p\left( X^{(test)}\right)$ are uniform distribution with Gaussian noise; $p \left( X \vert Y \right)$ is a deterministic function. \label{fig:data}}
	\end{centering}
\end{figure*}

We conducted experiments on elaborate synthetic data to validate and illustrate the following results:
(i) confirming the ability of SF and PSF to reduce the distance between training and test data; (ii) comparing SF and PSF against other well-known algorithms from the machine learning literature; (iii) highlighting that the success of these algorithms largely depends on whether the conditional condition for successful CSA is met by the data.\\

We generated four different data sets: \emph{radial} and \emph{periodic}, satisfying respectively the conditions of SF and PSF; \emph{smooth} and \emph{diagonal}, meeting respectively the assumptions of IW and SSA.
Figure \ref{fig:data} illustrates the four data sets (see Appendix \ref{app:Synthetic} for more details on data generation).

Data from the training distribution $\pxtr$ constitute the training data set $\Xtr$, while data from the test distribution $\pxte$ are evenly partitioned into a target data subset $\Xta$ and a test data subset $\Xte$. The training data $\Xtr$ and the target data $\Xta$ are used to train the CSA algorithm and the classification module, while the test data $\Xte$ is used only for performance evaluation.\\

All the data are processed using the CSA classification systems described above. The values chosen for the hyper-parameters of the different classification systems are detailed in Appendix \ref{app:Synthetic}.\\

In order to validate that SF and PSF meet the marginal CSA condition, we estimate the distance between the pdfs of the training and the test data before and after CSA using the Maximum Mean Discrepancy (MMD) distance \citep{Gretton2012}, and we report the \emph{percentage difference} in the distance; negative values denote a decrease in the distance between the marginal distributions. 

To validate that the effectiveness of CSA algorithms for classification depends on the satisfaction of their specific conditional CSA conditions, we employ a classification task and we evaluate the percentage difference in accuracy with and without CSA. The use of a relative measure \citep[analogous to percentage drop,][]{Torralba2011} allows us to account for the differences between the CSA systems (such as in the use of labels for adaptation, type of classifiers and implementation details). For SSA+SVM, we report a single result because of its deterministic nature; for the others algorithms involving a random initialization, we report mean and standard error estimated over 10 independent trials.\\

Table \ref{tab:mmd_changes} lists the percentage difference in the MMD distance between the marginal pdfs of the training and test data following the adoption of CSA. These results confirm that both SF and PSF are able to significantly reduce the distance between the marginal pdfs of the training and the test data, thus satisfying the marginal CSA condition.
\begin{table}
	\caption{Percentage difference in MMD distance between the training and the test distribution.
		$-100\%$ indicates a reduction of the MMD distance of two magnitude orders after applying SF or PSF. \label{tab:mmd_changes}}
	\begin{centering}
		\begin{tabular}{ccc}
			\hline
			{} & \textbf{\emph{\footnotesize{SF}}} & \textbf{\emph{\footnotesize{PSF}}}\tabularnewline
			\hline
			\textbf{\emph{\footnotesize{Radial}}} & $-100.1\% \pm 0.04\%$ & $-50.5 \pm 10.2\%$  \tabularnewline
			\textbf{\emph{\footnotesize{Periodic}}} & $-99.7\% \pm 0.01\%$ & $-87.3\%\pm 5.1\%$  \tabularnewline
			\textbf{\emph{\footnotesize{Smooth}}} & $-89.8\% \pm 2.3\%$ & $-88.2\% \pm 3.7\%$  \tabularnewline
			\textbf{\emph{\footnotesize{Diagonal}}} & $-80.3\% \pm 4.5\%$ & $-84.6\% \pm 4.6\%$  \tabularnewline
			\hline
			
		\end{tabular}
		\par\end{centering}

\end{table}

\begin{table*}
	\caption{Accuracy change when using different CSA systems on the four synthetic data sets. Accuracy without the CSA algorithm to left of the arrow and accuracy with the CSA algorithm to the right of the arrow. \label{tab:accuracy}}
	\begin{centering}
		\begin{tabular}{cccc}
			\hline
			{} & 
			\textbf{\emph{\footnotesize{SF+SVM}}} & 
			\textbf{\emph{\footnotesize{PSF+SVM}}} &
			\textbf{\emph{\footnotesize{IW+LSPC}}} 
			\tabularnewline
			\hline
			\textbf{\emph{\footnotesize{Radial}}} & $0.342\rightarrow0.779 \pm 0.03$ & $0.342\rightarrow0.479 \pm 0.05$ & $0.336\rightarrow0.598 \pm 0.06$   \tabularnewline
			\textbf{\emph{\footnotesize{Periodic}}} & $0.488\rightarrow0.5 \pm 0.01$ & $0.488\rightarrow0.568 \pm 0.02$ & $0.488\rightarrow0.512 \pm 0.0$   \tabularnewline
			\textbf{\emph{\footnotesize{Smooth}}} & $0.894\rightarrow0.476 \pm 0.06$ & $0.894\rightarrow0.522 \pm 0.08$ & $0.865\rightarrow0.936 \pm 0.02$ \tabularnewline
			\textbf{\emph{\footnotesize{Diagonal}}} & $0.866\rightarrow0.584 \pm 0.02$ & $0.866\rightarrow0.609 \pm 0.09$ & $0.768\rightarrow0.786 \pm 0.01$  \tabularnewline
			\hline
			
			{} & 
			\textbf{\emph{\footnotesize{SSA+SVM}}} &
			\textbf{\emph{\footnotesize{DAE+SVM}}} &
			\tabularnewline
			\hline
			\textbf{\emph{\footnotesize{Radial}}} & $0.342\rightarrow0.342$ & $0.342\rightarrow0.387 \pm 0.05$  \tabularnewline
			\textbf{\emph{\footnotesize{Periodic}}} & $0.488 \rightarrow 0.488$ & $0.488\rightarrow0.512 \pm 0.0$  \tabularnewline
			\textbf{\emph{\footnotesize{Smooth}}} & $0.89\rightarrow0.89$ & $0.894\rightarrow0.613 \pm 0.02$ \tabularnewline
			\textbf{\emph{\footnotesize{Diagonal}}} & $0.86\rightarrow0.932$ & $0.866\rightarrow0.514 \pm 0.0$ \tabularnewline
			\hline
			
		\end{tabular}
		\par\end{centering}
\end{table*}

\begin{table*}
	\caption{Percentage change in accuracy when using different CSA systems on the four synthetic data sets. \label{tab:percentage_increase}}
	\begin{centering}
		\begin{tabular}{cccc}
			\hline
			{} & 
			\textbf{\emph{\footnotesize{SF+SVM}}} & 
			\textbf{\emph{\footnotesize{PSF+SVM}}} &
			\textbf{\emph{\footnotesize{IW+LSPC}}}
			\tabularnewline 
			\hline
			\textbf{\emph{\footnotesize{Radial}}} & $\mathbf{+127.8\% \pm 9.4\%}$ & $+40.1\% \pm 14.7\%$ & $+7.8 \pm 2.1\%$  \tabularnewline
			\textbf{\emph{\footnotesize{Periodic}}} & $+0.33\% \pm 0.7\%$ & $\mathbf{+16.35\% \pm 5.0\%}$ & $+4.91\%\pm 0\%$ \tabularnewline
			\textbf{\emph{\footnotesize{Smooth}}} & $-46.76\% \pm 7.0\%$ & $-41.61\% \pm 9.0\%$ & $\mathbf{+8.36\% \pm 1.3\%}$ \tabularnewline
			\textbf{\emph{\footnotesize{Diagonal}}} & $-32.54\% \pm 2.4\%$ & $-29.63\% \pm 10.0\%$ & $+2.74\% \pm 0.7\%$ \tabularnewline
			\hline
			{} & 
			\textbf{\emph{\footnotesize{SSA+SVM}}} &
			\textbf{\emph{\footnotesize{DAE+SVM}}} &
			\tabularnewline
			\hline
			\textbf{\emph{\footnotesize{Radial}}} & $0\%$ & $+13.16 \pm 14.1\%$  \tabularnewline
			\textbf{\emph{\footnotesize{Periodic}}} & $0\%$ & $+4.92 \pm 0.0\%$  \tabularnewline
			\textbf{\emph{\footnotesize{Smooth}}} & $0\%$ & $-31.45 \pm 2.6\%$  \tabularnewline
			\textbf{\emph{\footnotesize{Diagonal}}} & $\mathbf{+8\%}$ & $-40.65 \pm 0.0\%$  \tabularnewline
			\hline

		\end{tabular}
		\par\end{centering}
\end{table*}

\rev{I added the table with raw performances. This table reports mean and standard error and allows for a more precise comparison of the results.}{Table} \ref{tab:accuracy} reports the raw accuracy of all the CSA classification systems before and after introducing CSA. The results shows that, even if the baselines of the five CSA classification systems are not exactly the same, they are still comparable; therefore, it makes sense to compare the percentage change in accuracy when processing a given data set.

Table \ref{tab:percentage_increase} reports the percentage difference in accuracy for all the CSA systems on the four data sets. These results highlight clearly a direct correlation between the success in classification and the satisfaction of the conditional CSA conditions related to the assumptions on which each algorithm works.
Indeed, the most relevant positive results in Table \ref{tab:percentage_increase} are obtained when a CSA classification system is used on data matching its assumptions. SF provides by far the best improvement on the \emph{radial} data set; as shown in Table \ref{tab:accuracy}, \rev{Referencing the table we added we can now argue that the performance of PSF is relevant because it raises performance above chance level.}{PSF is the only algorithm consistently increasing the performance above chance level} on the \emph{periodic} data set; IW and SSA offers the best improvements on the \emph{smooth} and the \emph{diagonal} data set respectively.

Consistently, violation of the assumptions results in very limited improvement or even decrease in classification  accuracy. 
For instance, SF yields a negligible increase in accuracy when applied to the \emph{periodic} data set, due to the fact that classification with a linear SVM remains basically at a random guess level before and after CSA, but it causes a severe drop in accuracy when applied to other data sets, due to the assumption mismatch. 
In contrast, PSF yields a significant improvement even when applied to the \emph{radial} data set, due to the fact that it is able to extrapolate a periodic structure under the data; however, it no longer works on the \emph{smooth} and the \emph{diagonal} data sets because probably it can not reconstruct a periodic structure from samples coming from a single period.
In general, PSF exhibits also a high variance, suggesting that, despite the supervised guidance, the algorithm is very sensitive to its initialization; different initial weight settings may lead PSF to try to extract very different periodic structures.
Finally, IW and SSA behave in a more conservative way, providing small or no improvements when their assumptions are violated, while DAE works only on the \emph{radial} and \emph{periodic} data sets but severely reduces the accuracy on the other data sets. 

This comparative study also suggests that there may be a trade-off between the percentage change in performance and the strictness of the assumptions: the stricter the assumptions, the higher the percentage change, positive if the conditions are met or negative if the conditions are not met, as exemplified by SF in Table \ref{tab:percentage_increase}. On the other hand, if the assumptions are looser, the variation in performance is limited between the case when the conditions are met and the case when the conditions are not met, as illustrated by IW in Table \ref{tab:percentage_increase}. This phenomenon may be well explained in the terms of the no-free lunch theorem \citep{Wolpert1997}.

\subsection{Real-World Data Set \label{sec:Real-World-Data-Experiments}}

We further carried out experiments on real-world data to validate the following statements: (i) a classification system using PSF can provide a statistically significant improvement over the baseline system without CSA, and (ii) PSF can provide competitive performances against other CSA algorithms reviewed in Section \ref{sec:BG_CSA}. \\

Real-world data are often very complex and do not perfectly fit the simple assumptions of the CSA algorithms. In this simulation, we chose emotional speech recognition (ESR) data sets to evaluate PSF for two reasons: (i) it is well known that ESR data sets are affected by covariate shift \citep{Schuller2010}, and (ii) ESR data is user-dependent and hence it may be modeled according to the assumption of PSF, so that each user could be specified by a different pdf $\PDF{p}{X^{user_i}}$ on a specific sub-domain $\domx^{user_i}$, while the conditional distribution of the emotional labels may be approximately the same for all the users $\pygx$.

In the following experiments four well-known ESR data sets are employed: the Berlin Emotional Database (EMODB) \citep{Burkhardt2005}, the Danish Emotional Speech Database (DES) \citep{Engberg2007}, Vera am Mittag (VAM) \citep{Grimm2008} and eNTERFACE (eNT) \citep{Martin2006}. This collection of data sets is very heterogeneous, containing recordings from different speakers, in different languages, with different labels and collected with different protocols (see Table \ref{tab:ESR-datasets} for details). \\

\begin{table*}
	\caption{Comparison of ESR data sets.
		\emph{\#Pos Val} and \emph{\#Neg Val} refer respectively to the number of samples with positive valence and negative valence. \label{tab:ESR-datasets}}
	\begin{centering}
		\begin{tabular}{ccccc}
			\hline
			{} & 
			\textbf{\emph{\footnotesize{\#Speakers}}} &
			\textbf{\emph{\footnotesize{Language}}} &
			\textbf{\emph{\footnotesize{Recording}}} &
			\textbf{\emph{\footnotesize{Labelling}}}  
			\tabularnewline 
			\hline
			\textbf{\emph{\footnotesize{EMODB}}} & 10 & German & acted & discrete (7 classes) \tabularnewline
			\textbf{\emph{\footnotesize{DES}}} & 4 & Danish & acted & discrete (5 classes) \tabularnewline
			\textbf{\emph{\footnotesize{VAM}}} & 47 & German & natural & continuous (3 dimensions) \tabularnewline
			\textbf{\emph{\footnotesize{eNT}}} & 43 & English & induced & discrete (6 classes) \tabularnewline
			\hline
			
			{} & 
			\textbf{\emph{\footnotesize{\#Samples (1-sec)}}} &
			\textbf{\emph{\footnotesize{\#Pos Val}}} &
			\textbf{\emph{\footnotesize{\#Neg Val}}} 
			\tabularnewline 
			\hline
			\textbf{\emph{\footnotesize{EMODB}}} & 1211 & 289 & 922   \tabularnewline
			\textbf{\emph{\footnotesize{DES}}} & 974 & 579 & 395   \tabularnewline
			\textbf{\emph{\footnotesize{VAM}}} & 2495 & 167 & 2328  \tabularnewline
			\textbf{\emph{\footnotesize{eNT}}} & 2988 & 877 & 2111  \tabularnewline
			\hline

		\end{tabular}
		\par\end{centering}
\end{table*}

All the recordings are pre-processed into one-second segments and converted into 72-dimensional feature vectors based on Mel-frequency cepstrum coefficients (MFCC) \citep{Eyben2010a}. 
All the labels are aligned using a standard procedure that reduces the heterogeneous labels to binary labels denoting a state of positive or negative valence (see Appendix \ref{app:Real-World} for details on feature extraction and label alignment).

We decided to use only speakers from EMODB, DES and eNTERFACE when performing CSA, excluding VAM because of the high unbalance between the valence classes (see Table \ref{tab:ESR-datasets}). We partition the data using the following protocol: in each trial, one speaker from either EMODB, DES or eNTERFACE is randomly selected; half of the samples from the selected speaker constitute the target set $\Xta$, while the remaining half constitutes the test set $\Xte$; all the samples from the three remaining data sets constitute the training set $\Xtr$. Adaptation is performed on the training and target data; classification uses training data for learning, target data for model selection and test data for evaluation. This protocol has two advantages: (i) in line with the most challenging scenarios in literature, this is a dataset-out and speaker-out scenario, in which the training set does not contain samples from the same data set or the same speaker in the test set, and (ii) preserving part of the samples only for test allows us to properly evaluate the degree of inductive generalization.\\

\rev{Here I provided more details about CV in general, and about SF and PSF in particular. Now all that is left in the appendix are just the numbers and the tables; no explanation or detail is left in the appendix. For readability, I left the tables in the appendix}{All the data are processed using the same classification systems used in the previous experiment. Hyper-parameters for the five classification systems are selected by cross-validation using a standard grid search method.} Details about the grid space for the hyper-parameters and the actual values selected through cross-validation are provided in Appendix \ref{app:Real-World}.

Moreover, to decrease the number of hyper-parameters, to reduce the computational time and to improve the results, we implemented an early stopping criterion for SF and PSF relying on the Kolmogorov-Smirnov (KS) test. The KS test is a statistical test assessing whether two univariate distributions are the same. Following \citet{Hassan2013}, we apply this test to pair of features to derive a gross estimation of the distance\footnote{This estimation is efficient, but clearly inaccurate as the distance between all the univariate distributions of $\PDF{p}{X_{1,i}}$ and $\PDF{p}{X_{2,i}}$ is not the same as the distance between the multivariate distributions $\PDF{p}{X_{1}}$ and $\PDF{p}{X_{2}}$.} between the distribution of the training $\mathbf{X^{tr}}$ and target data $\mathbf{X^{tar}}$. Given two samples $\mathbf{x}_{1}^\mathbf{tr}$ and $\mathbf{x}_{2}^\mathbf{tar}$ we apply the KS test feature-by-feature $KS\left(x_{1,j}^\mathbf{tr},x_{2,j}^\mathbf{tar}\right)$, and we average over all the feature $\mathbb{E}_j\left[KS\left(x_{1,j}^\mathbf{tr},x_{2,j}^\mathbf{tar}\right)\right]$. We stop training when the learned distribution of the training $\mathbf{Z^{tr}}$ and target data $\mathbf{Z^{tar}}$ achieve a minimum in the KS distance over the first 50 iterations.\\

In order to evaluate the performance in this unbalanced classification problem we employ the unweighed average recall (UAR): $\frac{1}{C}\sum_{c=1}^{C}recall(c)$, where $recall(c)$ denotes the recall for class $c$, computed as the number of correctly identified instances of class $c$ divided by all the instances of class $c$ \citep{Batliner2010}. 
For each configuration, we report the mean and the standard error achieved over 10 independent simulations.
In order to validate the contribution of PSF to classification, we perform a paired Wilcoxon test with the null hypothesis that the classification performance with and without CSA has the same mean (under the assumption that the results are symmetrically distributed around the true mean performance). Statistics for the hypothesis test are collected from 100 independent trials.\\

\begin{figure}
	\begin{centering}
		\includegraphics[scale=0.5]{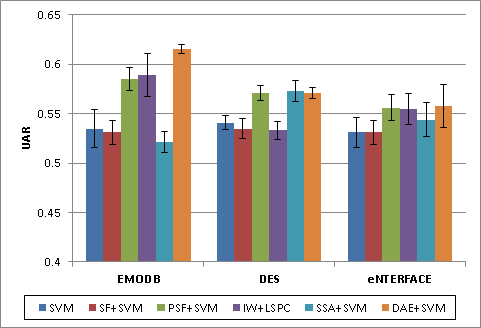}
	\par\end{centering}
	\caption{UAR Accuracies of different CSA methods along with the baselines. \label{fig:UAR}}
\end{figure}

Figure \ref{fig:UAR} shows the UAR of different classification systems on the three data sets using the protocol described above. After running the Wilcoxon test, we could reject the null hypothesis that classification with and without PSF is equivalent in the case of EMODB and eNTERFACE ($p$-value, respectively, $9.5\cdot10^{-5}$ and $6.9\cdot10^{-4}$), but not in the case of DES (at $p$-value $0.05$). Thus, the statistical test implies that PSF was indeed able to capture some relevant periodic structure in the conditional distribution when applied to the EMODB and eNTERFACE data sets. 
In general, the experimental results suggest that PSF is able to provide performances better or comparable to those of the other CSA algorithms, even if it is outperformed by DAE on a specific data set (EMODB).
%On the other hand, the negative test result on DES suggests that the better performance may have been due a random effect.
On the other hand, the SF algorithm failed to improve over the baseline. This was not surprising, as it can be easily explained by the fact that real-world data sets are too complex to comply with the tight assumption of a radial data structure.
Of the other CSA algorithms, DAE provided the best results with the highest UAR on EMODB and comparable UAR to other classification systems on the remaining data sets. Interestingly, IW yields good results on EMODB and eNTERFACE, but failed to provide an improvement on DES, potentially because DES speaker data may lie in sub-domains removed from the other data sets. SSA, instead, performed well on DES and eNTERFACE, but performed the worst on EMODB; this could hint at the fact that the PCA components of the EMODB test speaker data are not easily projected on the PCA components of all the other speakers.
%Interestingly, even when negative, the results of the CSA algorithms on individual data sets may allow us to get an insight on the structure of the conditional distribution; indeed, the failure of a CSA may be explained by the fact that a specific data set does not meet the assumptions of the CSA algorithm. For instance, on EMOBD, SSA algorithm fails to provide a significant improvement, . Similarly, on DES, the failure of IW could point to the fact that the pdfs describing the DES speakers lie in sub-domains removed from the other datasets. It is important, though, to remember that the failure of a CSA algorithm may be due to other factors, and that this explanation in terms of unmet assumptions is a likely hypothesis that however requires further validation.  

\section{Conclusions \label{sec:Conclusions}}

In this paper, we have studied the use of SF-like algorithms for CSA. We have rigorously showed that SF can perform CSA under the strict condition of a conditional distribution $\pygx$ explained by cosine neighbourhoodness. To overcome this limitation, we have then proposed the new PSF algorithm, which is able to perform CSA under looser and more realistic conditions on $\pygx$; indeed, while SF works only when the conditional distribution has a fixed radial structure, PSF can perform CSA adapting to data sets having different forms of periodic structure. We have demonstrated the strengths and limitations of SF and PSF on both synthetic and real-world data and we have compared their performances with other standard CSA algorithms. Experimental results have confirmed our theoretical conclusions and showed that PSF could be a competitive and efficient candidate for CSA, particularly when the structure of the conditional distribution is periodic. Moreover, our simulations have showed that by making the conditions for CSA explicit, we can not only predict when a CSA algorithm is likely to work, but we can also gain useful insights into the structure of the data even if a CSA algorithm were to fail.\\ 

Nevertheless, some non-trivial yet interesting points are left out in our current work and will be investigated in our ongoing research. 
From a theoretical viewpoint, we have not looked into the \emph{model misspecification assumption} discussed in \citet{Shimodaira2000}. The model misspecification assumption refers to a situation when the model employed to explain the data does not perfectly fit the data; this is normally due to the learning of a model from a family of simple parametric models that does not include the true model that generated the data \citep{Quinonero-Candela2009}. This situation may be explicitly stated for the algorithms in our simulations, too. Thus, future work would be directed at the model misspecification assumption in order to study theoretical bounds on the empirical risk when using SF and PSF.
From a practical viewpoint, it is also interesting to study the extension of our algorithms to the \emph{semi-supervised setting} \citep{Chapelle2006}. While we used SF and PSF in a purely unsupervised or supervised setting, our PSF has a potential to work in a semi-supervised way by processing unlabelled target samples alongside labelled source samples. In future work, we would study the dynamics of PSF in a semi-supervised scenario and compare it against other semi-supervised algorithms designed to perform CSA \citep{Margolis2011,Jiang2008}.

%Some limitations of this work may be the object of further research. In our discussion we did not explicit consider the assumption of \emph{model misspecification}, which justifies the adoption of CSA algorithms in \citet{Shimodaira2000}. The model misspecification assumption states that the model employed to explain the data can not perfectly fit the data; this is normally due to the learning of a model from a family of simple parametric models that does not contain the true model that generated the data \citep{Quinonero-Candela2009}. While this assumption tacitly lies behind the algorithms in our simulations, future work may be directed at making this assumption explicit and study theoretical bounds on the empirical risk that may be derived from it.

%On a practical level, it may also be interesting to study the extension of our algorithms to the \emph{semi-supervised setting} \citep{Chapelle2006}. While we ran SF and PSF in a purely unsupervised or supervised fashion, PSF has the potential to work in a semi-supervised way, too; indeed, we defined the PSF algorithm in such a way that it could straightforwardly process unlabelled samples alongside labelled samples. Future work may be devoted to analyse the dynamics of PSF in a semi-supervised scenario and to compare it against other semi-supervised algorithms designed to perform CSA \citep{Margolis2011,Jiang2008}.\\

Our results provide a deep understanding of the potential of SF and PSF in performing CSA. This understanding may be generalized to feature distribution learning algorithms and it could be further exploited to develop novel feature distribution learning algorithms with high computational efficiency. Reasoning in terms of the marginal and conditional CSA conditions, it may be possible to design new SF-like algorithms making explicit assumptions about the conditional distribution of the data. Given prior knowledge about the structure of a data set, this knowledge may be directly integrated to develop a SF variant tailored to the data; without explicit prior knowledge, metric learning algorithms could instead be employed or integrated in SF-based algorithms to indirectly learn the conditional structure of the data.

\begin{acknowledgements}
F. M. Zennaro's work was supported by the Kilburn PhD studentship.
\end{acknowledgements}

\appendix

\newcommand{\fI}{f_{A1}}
\newcommand{\fII}{f_{A2}}
\newcommand{\fIII}{f_{A3}}
\newcommand{\fIV}{f_{A4}}
\newcommand{\fV}{f_{A5}}
\newcommand{\fIfIV}{f_{A1:A4}}
\newcommand{\fIIfIV}{f_{A2:A4}}
\newcommand{\fIfII}{f_{A1:A2}}

\newcommand{\xbias}{\mathbf{b}}
\newcommand{\xbiasj}{b_j}
\newcommand{\xdispl}{\bar{\mathbf{x}}}
\newcommand{\xdisplj}{\bar{x}_j}
\newcommand{\fil}{\DATAMATRIXelem{f}{i}{l}}
\newcommand{\fonel}{\DATAMATRIXelem{f}{1}{l}}
\newcommand{\ftwol}{\DATAMATRIXelem{f}{2}{l}}
\newcommand{\wjl}{\DATAMATRIXelem{w}{j}{l}}
\newcommand{\fdispl}{\bar{\mathbf{f}}}
\newcommand{\fdisplj}{\bar{f}_j}
\newcommand{\fdispll}{\bar{f}_l}
\newcommand{\ftildeonel}{\DATAMATRIXelem{\tilde{f}}{1}{l}}
\newcommand{\ftildetwol}{\DATAMATRIXelem{\tilde{f}}{2}{l}}
\newcommand{\ftildedispl}{\bar{\tilde{\mathbf{f}}}}
\newcommand{\ftildedisplj}{\bar{\tilde{f}}_j}
\newcommand{\ftildedispll}{\bar{\tilde{f}}_l}
\newcommand{\fhatdispl}{\bar{\mathbf{\hat{f}}}}
\newcommand{\fhatdisplj}{\bar{\hat{f}}_j}
\newcommand{\zonel}{\DATAMATRIXelem{z}{1}{l}}
\newcommand{\ztwol}{\DATAMATRIXelem{z}{2}{l}}
\newcommand{\zdispl}{\bar{\mathbf{z}}}
\newcommand{\zdisplj}{\bar{z}_j}
\newcommand{\zdispll}{\bar{z}_l}
\newcommand{\cc}{\mathbf{c}}
\newcommand{\ccj}{c_j}
\newcommand{\ccl}{c_l}

\newcommand{\f}{\DATAMATRIXrow{f}{i}{}}
\newcommand{\ftilde}{\DATAMATRIXrow{\tilde{f}}{i}{}}
\newcommand{\fone}{\DATAMATRIXrow{f}{1}{}}
\newcommand{\fonej}{\DATAMATRIXelem{f}{1}{j}}
\newcommand{\ftwo}{\DATAMATRIXrow{f}{2}{}}
\newcommand{\ftwoj}{\DATAMATRIXelem{f}{2}{j}}
\newcommand{\ftildeone}{\DATAMATRIXrow{\tilde{f}}{1}{}}
\newcommand{\ftildeonej}{\DATAMATRIXelem{\tilde{f}}{1}{j}}
\newcommand{\ftildetwo}{\DATAMATRIXrow{\tilde{f}}{2}{}}
\newcommand{\ftildetwoj}{\DATAMATRIXelem{\tilde{f}}{2}{j}}
\newcommand{\fonehat}{\DATAMATRIXrow{\hat{f}}{1}{}}
\newcommand{\fonehatj}{\DATAMATRIXelem{\hat{f}}{1}{j}}
\newcommand{\fhattwo}{\DATAMATRIXrow{\hat{f}}{2}{}}
\newcommand{\fhattwoj}{\DATAMATRIXelem{\hat{f}}{2}{j}}
\newcommand{\zonej}{\DATAMATRIXelem{z}{1}{j}}
\newcommand{\ztwoj}{\DATAMATRIXelem{z}{2}{j}}
\newcommand{\lambdavectork}{\LABELVECTOR{\lambda}{k}}
\newcommand{\Zfixed}{\DATAMATRIX\DATA{\bar{Z}}{}}
\newcommand{\zfixedj}{\DATAMATRIXelem{z}{\emptyarg}{\bar{j}}}
\newcommand{\zifixedj}{\DATAMATRIXelem{z}{i}{\bar{j}}}
\newcommand{\rvZfixedj}{\RV{Z_{\emptyarg,\bar{j}}}{}}

\section{Theoretical Analysis of Sparse Filtering for Covariate Shift Adaptation \label{sec:Theory-SF}}

\subsection{Proof of Proposition 1 \label{app:Proof-of-Proposition-1}}

\begin{proposition}
	The sub-domain $\domz$ of the representations $\z$ learned by SF is $\closedinterval{0}{1}^\nZfeatures$.
\end{proposition}

\textbf{Proof.} Let $\xone \in \domx$ be a generic data sample to be processed through SF. Let the matrix $\Ftilde$
be the output of the $\lII$-normalization along the rows of the SF algorithm, that is:
\[
\Ftilde = \lIIrow{\absval{\W \X}}.
\]
The final output of the SF algorithm is then:
\[
\zone = \frac{\ftildeonej}{\sqrt{\sumjfeatZ \left( \ftildeonej \right)^2 }}.
\]
Every feature $\zonej$ is given by the feature value $\ftildeonej$ normalized by the $\lII$-norm of $\ftildeone$. Therefore, it follows that each feature $\zonej$ is bounded within $\closedinterval{0}{1}$. Consequently, the representation $\zone$ is bounded within the hyper-cube $\closedinterval{0}{1}^\nZfeatures$. $\QED$

\subsection{Proof of Proposition 2 \label{app:Proof-of-Proposition-2}}

\begin{proposition}
	For each learned feature $\zj$, the SF algorithm bounds $\mean{\rvZfeatj} \in \closedinterval{\epsilon}{1}$ and $\var{\rvZfeatj} \in \closedinterval{0}{1-\epsilon^2}$, where $\epsilon>0$ is an arbitrarily small value defined in the non-linearity of SF. 
	Moreover, if we make the assumption that learned representations are $\closedinterval{1}{k}$-sparse in population and lifetime, and that $\epsilon$ is negligible, then we have the bounds $\mean{\rvZfeatj} \in \closedinterval{\frac{1}{\nXsamples}}{\frac{k}{\nXsamples}}$ and $\var{\rvZfeatj} \in \closedinterval{\frac{\nXsamples-k^{2}}{\nXsamples^{2}}}{\frac{\nXsamples k-1}{\nXsamples^{2}}}$.
\end{proposition}

\textbf{Proof.} The proof of this proposition is based on the following logical steps: (a) re-statement of the basic properties of the learned representations; (b) estimation of the expected value of the distribution of a learned feature (with and without the assumption of $k$-sparsity); (c) estimation of the second moment of the distribution of a learned feature (with and without the assumption of $k$-sparsity); (d) estimation of the variance of the distribution of a learned feature (with and without the assumption of $k$-sparsity).

(a) Let us consider the $\lII$-normalization steps defining $\Ftilde$ and $\Z$. These transformations have two main effects: they constrain all the values in $\Ftilde$ and $\Z$ to be within $\closedinterval{0}{1}$; and, they force features or samples to have a square total activation of 1. Formally:
\[
1 \leq \ftildeij \leq 0, \,\, 
1 \leq \zij \leq 0, \,\,\, 
\forall 
1\leq j \leq \nZfeatures, \,
1\leq i \leq \nXsamples,
\]
\[ \sumisamples\left( \ftildeij \right)^2=1,\,\,\,1\leq j\leq L \]
\[  \sumjfeatZ\left( \zij \right)^2=1,\,\,\, 1\leq i\leq N. \]

(b) Let us now consider a given feature and, for clarity, let us denote this fixed feature as $\bar{j}$ to underline the fact that it is not going to change in the following analysis. We can now analyse the distribution of $\zfixedj$ by considering its main statistical moments. Let us start by analysing the expected value of the random variable $\rvZfixedj$:
\[
\mean{\rvZfixedj} \estimate \frac{1}{\nXsamples} \sumisamples \zifixedj.
\]
After normalizing along the column, the quantity $\sumisamples \zifixedj$ is not rigidly constrained. The value of the feature $\zifixedj$ can range between $\epsilon$ (if the feature $\bar{j}$ happens to be inactive for the sample $i$) and $1$ (if the feature is the only active feature for the sample $i$). Therefore, the expected value can be bound in:
\[
\epsilon \leq \mean{\rvZfixedj} \leq 1.
\]
Let us now make the assumption that the feature $\zfixedj$ is at most $k$-sparse, with $1 < k < \nZfeatures$, that is, it is active on a number $k$ of samples, with $k$ greater than $1$ and smaller than $\nZfeatures$. This assumption is justified by considering the properties of population sparsity and lifetime sparsity of SF \citep{Ngiam2011}. In this case, the expected value can be bound in:
\[
\frac{1+(\nXsamples-1)\epsilon}{\nXsamples} \leq \mean{\rvZfixedj} \leq \frac{k+(\nXsamples-k)\epsilon}{\nXsamples}.
\]
Moreover, if we assume that $\epsilon$ is negligible, then the final bound can be re-written as:
\[
\frac{1}{\nXsamples} \leq \mean{\rvZfixedj} \leq \frac{k}{\nXsamples}.
\]
This proves the first part of our statement.

(c) Let us now consider the estimation of the second moment:
\[
\mom{2}{\rvZfixedj} = \mean{\left(\rvZfixedj\right)^2}
\estimate \frac{1}{\nXsamples}\sumisamples\zifixedj^2.
\]
For the same reason we gave above about the admissible values for the feature $\zfixedj$, the second moment can be bound in:
\[
\epsilon^2 \leq \mom{2}{\rvZfixedj} \leq 1.
\]
Under the assumption of $k$-sparsity of $\zfixedj$, we can get the tighter bounds:
\[
\frac{1+(\nXsamples-1)\epsilon^2}{\nXsamples} \leq \mom{2}{\rvZfixedj} \leq \frac{k+(\nXsamples-k)\epsilon^2}{\nXsamples}.
\]
(d) Finally, let us consider the estimation of the variance of $\rvZfixedj$:
\[ 
\var{\rvZfixedj} = \mean{\rvZfixedj{}^2} - \mean{\rvZfixedj}^2. 
\] 
Again, using the values we computed for the second moment and the expected value, we can define the following bounds for the variance:
\begin{eqnarray*}
	\epsilon^{2}-1^{2} & \leq \var{\rvZfixedj} \leq & 1-\epsilon^{2}\\
	0 & \leq \var{\rvZfixedj} \leq & 1-\epsilon^{2}.
\end{eqnarray*}
Under the assumption of $k$-sparsity we can recompute the bounds: 
\begin{eqnarray*}
	\var{\rvZfixedj} & \geq & \frac{1+(\nXsamples-1)\epsilon^{2}}{\nXsamples}-\left(\frac{k+(\nXsamples-k)\epsilon}{\nXsamples}\right)^{2}\\
	& \geq & \frac{\nXsamples(1-\epsilon^{2}+2k\epsilon^{2}-2k\epsilon)-k^{2}(1+\epsilon^{2}-2\epsilon)}{\nXsamples^{2}},
\end{eqnarray*}
\begin{eqnarray*}
	\var{\rvZfixedj} & \leq & \frac{k+(\nXsamples-k)\epsilon^{2}}{\nXsamples}-\left(\frac{1+(\nXsamples-1)\epsilon}{\nXsamples}\right)^{2}\\
	& \leq & 
	\frac{\nXsamples(k-k\epsilon^{2}+2\epsilon^{2}-2\epsilon)-1-\epsilon^{2}+2\epsilon}{\nXsamples^{2}}.
\end{eqnarray*}
If we take $\epsilon$ to be negligible, then the bounds of the variance are:
\[
\frac{\nXsamples-k^{2}}{\nXsamples^{2}} \leq \var{\rvZfixedj} \leq \frac{\nXsamples k-1}{\nXsamples^{2}}.
\]
This proves the last part of our statement. $\QED$

\subsection{Proof of Theorem on the Preservation of the Cosine Neighbourhoodness in SF} \label{app:preservation-cosine-neigh}

\begin{theorem} 
	Let $\xone,\xtwo \in \origspace$ be two original data samples and let $\zone,\ztwo \in \learnspace$ be their representations computed by SF. If the cosine distance between the original samples is arbitrarily small $\dist{C}{\xone}{\xtwo}<\delta$, for $\delta>0$, then the Euclidean distance between the computed representations is arbitrarily small $\dist{E}{\zone}{\ztwo}<\epsilon$, for $\epsilon>0$, and $\epsilon = \nZfeatures \multiplication \left(\frac{k + \absval{\sqrt{2\delta-\delta^{2}}}} {\APPLIEDFUNCTION{\lII}{\ftildetwo}} -
	\frac{1}{\APPLIEDFUNCTION{\lII}{\ftildeone}}\right)$,
	where $k$ is a constant accounting for partial collinearity and $\APPLIEDFUNCTION{\lII}{\ftilde}$ is the $\lII$-norm of the representations computed by SF after step A3.
\end{theorem}

\textbf{Proof.} In order to prove this theorem we adopt the following strategy: we compute the representations at each step of the computation (before SF, after steps A1 and A2, after step A3 and after step A4) and we upper bound the displacement accounting for the Euclidean distance between the representations.

Recall that given two generic points $\xone$ and $\xtwo$, we can express $\xtwo$ as a function of $\xone$ plus a \emph{displacement} vector $\xdispl$:
\begin{equation}
\xtwo=\xone+\xdispl,\label{eq:displacement0}
\end{equation}
so that we can easily account for the Euclidean distance between $\xone$ and $\xtwo$ just as a function of the displacement vector $\xdispl$:
\[
\dist{E}{\xone}{\xtwo} = \APPLIEDFUNCTION{\lII}{\xdispl}.
\]

\textbf{(Before SF.)} Let us now consider two points $\xone$ and $\xtwo$ which are almost collinear with an arbitrary small cosine distance $\dist{C}{\xone}{\xtwo} < \delta$. We can then express $\xone$ as a point collinear with $\xtwo$ to which a \emph{bias} vector $\xbias$ is added:
\[
\xtwo=k \xone + \xbias,
\]
where $k \in \domR$, $k\neq0$ is a constant that preserves collinearity.
With no loss of generality, we will assume $k>1$; we exclude values of $k$ smaller than zero which would generate a reflection (reflections are not relevant for the following treatment as they induce a cosine distance far greater than $\delta$) and we ignore values of $k$ falling between zero and one (in such a case, our proof will hold once we swap $\xone$ and $\xtwo$). The bias vector $\xbias$ accounts for a relative displacement between the perfectly collinear sample $k \xone$ and the almost collinear sample $k \xone + \xbias$.

With reference to Equation \ref{eq:displacement0}, the displacement vector $\xdispl$ is:
\begin{equation}
\xdispl = (k-1) \xone + \xbias,\label{eq:displacement1}
\end{equation}
from which follows that:
\[
\dist{E}{\xone}{\xtwo} = \APPLIEDFUNCTION{\lII}{\xdispl} = \APPLIEDFUNCTION{\lII}{(k-1)\xone + \xbias}.
\]

\textbf{(Before SF - Upper bound)} To upper bound $\dist{E}{\xone}{\xtwo}$, we can evaluate the maximum value that $\APPLIEDFUNCTION{\lII}{\xdispl}$ can reach, consistent with the constraint of a bounded cosine distance $\dist{C}{\xone}{\xtwo}$. Formally, we set up the optimization problem:
\[
\argmaxproblem{\xdispl \in \origspace}{\APPLIEDFUNCTION{\lII}{\xdispl}},
\]
under the constraint:
\[
\dist{C}{\xone}{\xtwo}<\delta.
\]
The maximization can be rewritten as: 
\begin{eqnarray*}
	\argmaxproblem{\xdispl \in \origspace}{\APPLIEDFUNCTION{\lII}{\xdispl}} & = &
	\argmaxproblem{\xdisplj \in \domR}{\sqrt{\sumjfeatX \xdisplj^2}}\\
	& = &
	\argmaxproblem{\xbiasj \in \domR}{\sqrt{\sumjfeatX \left( (k-1)\xonej+\xbiasj \right)^2}}\\
	& = &
	\argmaxproblem{\xbiasj \in \domR}{\sqrt{\sumjfeatX \xbiasj^2}}\\
	& = & \argmaxproblem{\xbiasj \in \domR}{\xbiasj},
\end{eqnarray*}
assuming: (i) that $\xone$ and $k$ are given and fixed, and (ii) that $\xonej$ and $\xbiasj$ are both positive (as this constitutes the worst case that needs to be considered in the analysis of the upper bound). An upper bound on the displacement $\xdispl$ can be then computed from the solution to the individual constrained optimization problems for each component $\xbiasj$:
\[
\maxproblem{\xbiasj \in \domR}{\xbiasj},
\]
under the constraint:
\begin{eqnarray*}
	\delta & > & \dist{C}{\xone}{\xtwo}\\
	& = & \dist{C}{\xone}{k \xone + \xbias}.
\end{eqnarray*}
By construction, we know that $\dist{C}{\xone}{k \xone}=0$. Therefore the entire cosine distance must be accounted by the bias vector $\xbias$. Trigonometrically, from the cosine distance $\delta$ we can recover the angle opposite to a cathetus corresponding to the radius of an hypersphere centred on $k \xone$ and bounding the module of $\xbias$. Let $\theta$ be the underlying angle between $\xone$ and $\xtwo$:
\begin{eqnarray*}
	\delta & = & 1-\APPLIEDFUNCTION{\cos}{\theta}\\
	\theta & = & \APPLIEDFUNCTION{\arccos}{1-\delta}.
\end{eqnarray*}
The radius of the hypersphere centred on $k \xone$ inducing at most a cosine distance $\delta$ is:
\begin{eqnarray*}
	\xbiasj & \leq & \xonej \APPLIEDFUNCTION{\sin}{\APPLIEDFUNCTION{\arccos}{1-\delta}}\\
	& = & \xonej \sqrt{1-(1-\delta)^{2}}\\
	& = & \xonej \sqrt{2\delta-\delta^{2}}.
\end{eqnarray*}
Substituting this value in Equation \ref{eq:displacement1}, the displacement on each component can the be upper bounded as:
\begin{eqnarray*}
	\xdisplj & = & (k-1) \xonej + \xbiasj\\
	& \leq & (k-1) \xonej + \xonej \sqrt{2\delta-\delta^{2}}\\
	& = & \xonej \left(k-1+\sqrt{2\delta-\delta^{2}}\right).
\end{eqnarray*}
This upper bound depends on the original cosine distance $\delta$, but more significantly on the module of $\xone$ and the stretching constant $k$. Indeed, the Euclidean distance along each component is given by the stretch ($\xonej \left(k-1\right)$) plus a small distance due to the angle ($\xonej \sqrt{2\delta-\delta^{2}}$).

\textbf{(Steps A1 and A2)} Let us now apply the linear projection and the absolute-value function defined in transformation A1 and A2:
\begin{eqnarray*}
	\fone = \APPLIEDFUNCTION{\fIfII}{\xone} & = & \absval{\W\xone}\\
	\ftwo = \APPLIEDFUNCTION{\fIfII}{\xtwo} & = & \absval{\W \left(k\xone + \xbias\right)} = k\fone \pm \absval{\W \xbias}.
\end{eqnarray*}
Component-wise we have:
\begin{eqnarray*}
	\fonel & = & \absval{\sumjfeatX \wjl \xonej}\\
	\ftwol & = & k \fonel + \absval{\W \xbias}_l =
	k \absval{\sumjfeatX \wjl \xonej} \pm \absval{\sumjfeatX \wjl \xbiasj}.
\end{eqnarray*}
The new displacement and the new Euclidean distance are:
\begin{equation}
\fdispll = (k-1) \fonel \pm \absval{\W\xbias}_l \label{eq:displacement2}
\end{equation}
\[
\dist{E}{\fone}{\ftwo} = \APPLIEDFUNCTION{\lII}{\fdispl} = \sqrt{\sum_{l=1}^{L} \left( (k-1)\fonel \pm \absval{\W\xbias}_l \right)^2}.
\]

\textbf{(Steps A1 and A2 - Upper bound)} The upper bound of each component
of the new bias vector follows immediately:
\begin{eqnarray*}
	\absval{\W \xbias}_l & = & 
	\absval{\sumjfeatX \wjl \xbiasj}\\
	& \leq & \absval{\sumjfeatX \wjl \xonej \sqrt{2\delta-\delta^2}}\\
	& = & \absval{\sqrt{2\delta-\delta^2}} \absval{\sumjfeatX \wjl \xonej},
\end{eqnarray*}
and then the upper bound on each component of the displacement in Equation \ref{eq:displacement2} is:
\begin{eqnarray*}
	\fdispll & \leq &
	(k-1) \fonel + \absval{\sqrt{2\delta-\delta^2}} \absval{\sumjfeatX \wjl \xonej}\\
	& = & 
	(k-1) \absval{\sumjfeatX \wjl \xonej} + \absval{\sqrt{2\delta-\delta^2}} \absval{\sumjfeatX \wjl \xonej} \\
	& = &
	\left(k-1+\absval{\sqrt{2\delta-\delta^2}}\right) \absval{\sumjfeatX \wjl \xonej}.
\end{eqnarray*}

\textbf{(Step A3)} Let us now apply the normalization along the rows defined in transformation A3:
\begin{eqnarray*}
	\ftildeonel = \APPLIEDFUNCTION{\fIII}{\fonel} & = & \frac{\fonel}{\sqrt{\sumisamples \fil^2}}\\	
	\ftildetwol = \APPLIEDFUNCTION{\fIII}{\ftwol} & = & 
	\frac{k \fonel + \absval{\W\xbias}_l}{\sqrt{\sumisamples \fil^2}} =
	k\ftildeonel + \frac{\absval{\W\xbias}_l}{\sqrt{\sumisamples \fil^2}}.
\end{eqnarray*}
Notice that the denominator is given by a feature-dependent sum across $\nXsamples$ samples; for simplicity, we will take this value to be a constant $\set{\ccl}_{l=1}^{\nZfeatures}$, $\ccl \in \domR$:
\begin{eqnarray*}
	\ftildeonel & = & \frac{\fonel}{\ccl}\\
	\ftildetwol & = & k \ftildeonel + \frac{\absval{\W\xbias}_l}{\ccl}.
\end{eqnarray*}
The new displacement and the new Euclidean distance are:
\begin{equation}
\ftildedispll = (k-1) \ftildeonel + \frac{\absval{\W\xbias}_l}{\ccl} \label{eq:displacement3}
\end{equation}
\[
\dist{E}{\ftildeone}{\ftildetwo} = \APPLIEDFUNCTION{\lII}{\ftildedispl} =
\sqrt{\sum_{l=1}^{\nZfeatures} \left( (k-1) \ftildeonel + \frac{\absval{\W\xbias}_l}{\ccl} \right)^2}.
\]
\textbf{(Step A3 - Upper bound)} The upper bound of each component
of the new bias vector follows immediately:
\[
\frac{\absval{\W\xbias}_l}{\ccl} \leq
\frac{\absval{\sqrt{2\delta-\delta^2}} \absval{\sumjfeatX \wjl \xonej}}{\ccl},
\]
and then the upper bound on each component of the displacement in
Equation \ref{eq:displacement3} is:
\begin{eqnarray*}
	\ftildedispll & \leq &
	(k-1) \ftildeonel + \frac{\absval{\sqrt{2\delta-\delta^2}} \absval{\sumjfeatX \wjl \xonej}}{\ccl}\\
	& = &
	(k-1) \frac{\fonel}{\ccl} + \frac{\absval{\sqrt{2\delta-\delta^2}} \absval{\sumjfeatX \wjl \xonej}}{\ccl}\\
	& = &
	(k-1) \frac{\absval{\sumjfeatX \wjl \xonej}}{\ccl} + \frac{\absval{\sqrt{2\delta-\delta^2}} \absval{\sumjfeatX \wjl \xonej}}{\ccl}\\
	& = &
	\frac{k-1+\absval{\sqrt{2\delta-\delta^2}}}{\ccl} \absval{\sumjfeatX \wjl \xonej}\\
	& = &
	\frac{1}{\ccl} \fdispll.
\end{eqnarray*}
Not surprisingly, after transformation A3, the Euclidean distance $\dist{E}{\ftildeone}{\ftildetwo}$ is just rescaled since each component of the displacement $\fdispll$ is reduced by a factor $\frac{1}{\ccl} = \frac{1}{\sqrt{\sumisamples \fil^2}}$.

\textbf{(Step A4)} Finally, let us apply the normalization along the
samples defined in transformation A4:
\begin{eqnarray*}
	\zonel & = & \APPLIEDFUNCTION{\fIV}{\ftildeonel} = 
	\frac{\ftildeonel}{\APPLIEDFUNCTION{\lII}{\ftildeone}} = \frac{\frac{\fonel}{\ccl}}{\APPLIEDFUNCTION{\lII}{\ftildeone}}\\
	\ztwol & = & \APPLIEDFUNCTION{\fIV}{\ftildetwol} = 
	\frac{\ftildetwol}{\APPLIEDFUNCTION{\lII}{\ftildetwo}} = 
	\frac{k\ftildeonel + \frac{\absval{\W\xbias}_l}{\ccl}} {\APPLIEDFUNCTION{\lII}{\ftildetwo}} = \frac{k \frac{\fonel}{\ccl} }{\APPLIEDFUNCTION{\lII}{\ftildetwo}} + \frac{\frac{\absval{\W\xbias}_l}{\ccl}}{\APPLIEDFUNCTION{\lII}{\ftildetwo}}.
\end{eqnarray*}
Let us now consider the first term of $\ztwol$ and let us multiply it by $\frac{\APPLIEDFUNCTION{\lII}{\ftildeone}}{\APPLIEDFUNCTION{\lII}{\ftildeone}}$:
\[
\ztwol = \frac{k \frac{\fonel}{\ccl} }{\APPLIEDFUNCTION{\lII}{\ftildetwo}} \multiplication \frac{\APPLIEDFUNCTION{\lII}{\ftildeone}}{\APPLIEDFUNCTION{\lII}{\ftildeone}} + \frac{\frac{\absval{\W\xbias}_l}{\ccl}}{\APPLIEDFUNCTION{\lII}{\ftildetwo}} =
k\zonel \frac{\APPLIEDFUNCTION{\lII}{\ftildeone}}{\APPLIEDFUNCTION{\lII}{\ftildetwo}} + \frac{\frac{\absval{\W\xbias}_l}{\ccl}}{\APPLIEDFUNCTION{\lII}{\ftildetwo}}.
\]
The new displacement and the new Euclidean distance are:
\begin{equation}
\zdispll = \left( k \frac{\APPLIEDFUNCTION{\lII}{\ftildeone}}{\APPLIEDFUNCTION{\lII}{\ftildetwo}} - 1 \right) \zonel + \frac{\absval{\W\xbias}_l}{\ccl \APPLIEDFUNCTION{\lII}{\ftildetwo}} \label{eq:displacement4}
\end{equation}
\begin{equation}
\dist{E}{\zone}{\ztwo} = \APPLIEDFUNCTION{\lII}{\zdispl} =
\sqrt {\sum_{l=1}^{\nZfeatures} \left( \left( k \frac{\APPLIEDFUNCTION{\lII}{\ftildeone}}{\APPLIEDFUNCTION{\lII}{\ftildetwo}} - 1 \right) \zonel + \frac{\absval{\W\xbias}_l}{\ccl \APPLIEDFUNCTION{\lII}{\ftildetwo}} \right)^2}. \label{eq:distance}
\end{equation}
For consistency, notice that if $\xone$ and $\xtwo$ were to be collinear, then $\APPLIEDFUNCTION{\lII}{\ftildetwo} = k\APPLIEDFUNCTION{\lII}{\ftildeone}$, and, by construction, $\xbias=0$; therefore, in case of collinearity, $\dist{E}{\zone}{\ztwo}$  computed in Equation \ref{eq:distance} would be zero.

\textbf{(Step A4 - Upper bound)} Now, the upper bound of each component of the bias vector can be immediately evaluated:
\begin{eqnarray*} 
	\frac{\absval{\W\xbias}_l}{\ccl \APPLIEDFUNCTION{\lII}{\ftildetwo}} & \leq &
	\frac{\absval{\sqrt{2\delta-\delta^2}} \absval{\sumjfeatX \wjl \xonej}}{\ccl \APPLIEDFUNCTION{\lII}{\ftildetwo}},
\end{eqnarray*}
and then the upper bound on each component of the displacement:
\begin{eqnarray*}
	\zdispll & \leq &
	\left( k \frac{\APPLIEDFUNCTION{\lII}{\ftildeone}}{\APPLIEDFUNCTION{\lII}{\ftildetwo}} - 1 \right) \zonel +
	\frac{\absval{\sqrt{2\delta-\delta^2}} \absval{\sumjfeatX \wjl \xonej}}{\ccl \APPLIEDFUNCTION{\lII}{\ftildetwo}}\\
	& = & 
	\left( k \frac{\APPLIEDFUNCTION{\lII}{\ftildeone}}{\APPLIEDFUNCTION{\lII}{\ftildetwo}} - 1 \right) \frac{\fonel}{\ccl \APPLIEDFUNCTION{\lII}{\ftildeone}} +
	\frac{\absval{\sqrt{2\delta-\delta^2}} \absval{\sumjfeatX \wjl \xonej}}{\ccl \APPLIEDFUNCTION{\lII}{\ftildetwo}}\\
	& = &
	\left( k \frac{\APPLIEDFUNCTION{\lII}{\ftildeone}}{\APPLIEDFUNCTION{\lII}{\ftildetwo}} - 1 \right) \frac{\absval{\sumjfeatX \wjl \xonej}}{\ccl \APPLIEDFUNCTION{\lII}{\ftildeone}} +
	\frac{\absval{\sqrt{2\delta-\delta^2}} \absval{\sumjfeatX \wjl \xonej}}{\ccl \APPLIEDFUNCTION{\lII}{\ftildetwo}}\\
	& = &
	\frac{\absval{\sumjfeatX \wjl \xonej}}{\ccl} 
	\left( \frac{k + \absval{\sqrt{2\delta-\delta^2}}}{\APPLIEDFUNCTION{\lII}{\ftildetwo}} - \frac{1}{\APPLIEDFUNCTION{\lII}{\ftildeone}}  \right).
\end{eqnarray*}

Notice that $\frac{\absval{\sumjfeatX \wjl \xonej}}{\ccl}<1$ since $\ccl = \sqrt{\sumisamples \fil^2 }$. Thus:
\[
\zdispll \leq \left( \frac{k + \absval{\sqrt{2\delta-\delta^2}}}{\APPLIEDFUNCTION{\lII}{\ftildetwo}} - \frac{1}{\APPLIEDFUNCTION{\lII}{\ftildeone}}  \right).
\]
The overall Euclidean distance between the representations $\zone$ and $\ztwo$ can then be bounded by:
\begin{eqnarray*}
	\dist{E}{\zone}{\ztwo} & = &
	\sqrt{\sum_{l=1}^{\nZfeatures} \zdispll^2}\\
	& \leq &
	\nZfeatures \multiplication \left( \frac{k + \absval{\sqrt{2\delta-\delta^2}}}{\APPLIEDFUNCTION{\lII}{\ftildetwo}} - \frac{1}{\APPLIEDFUNCTION{\lII}{\ftildeone}}  \right).
\end{eqnarray*}
Thus $\epsilon=\nZfeatures \multiplication \left( \frac{k + \absval{\sqrt{2\delta-\delta^2}}}{\APPLIEDFUNCTION{\lII}{\ftildetwo}} - \frac{1}{\APPLIEDFUNCTION{\lII}{\ftildeone}}  \right)$.  $\QED$

\section{Periodic Sparse Filtering \label{sec:SF}}

\subsection{Pseudo-Code for the Derivative of PSF \label{app:Pseudo-Code-for-PSF}}

Algorithm \ref{alg} lists the pseudo-code for the gradient descent on the PSF loss function.

\begin{algorithm}[h] \caption{Derivative for PSF \label{alg}}
	\begin{algorithmic}[1] 
		\Statex \textbf{Input:} input data $\X$; weight matrix $\W$; PSF output $\Z$; PSF intermediate representations $\IntReprI$, $\F$, $\Ftilde$.
		\Statex \textbf{Hyper-params:} lambda vector $\lambdavector$.
		\Statex 
		\State $\vectorize{\partderiv{\Z}{\Ftilde}}_{i,j} 
		\leftarrow
		\frac
		{\sqrt{\sumjfeatZ \ftildeij^2} - \zij \multiplication \sumjfeatZ\ftildeij}
		{\sumjfeatZ\ftildeij^2}$
		\State $\vectorize{\partderiv{\Z}{\F}}_{i,j} 
		\leftarrow 
		\frac
		{\vectorize{\partderiv{\Z}{\Ftilde}}_{i,j} \sqrt{\sumisamples\fij^2} -
			\ftildeij \multiplication \sumisamples \left( \vectorize{\partderiv{\Z}{\Ftilde}}_{i,j} \multiplication \fij\right)}
		{\sumisamples\fij^2}$
		\State $\partderiv{\Z}{\IntReprI} \leftarrow \partderiv{\Z}{\F} \multiplication \cos \IntReprI$
		\State $\partderiv{\Z}{\W} \leftarrow \lambdavector \partderiv{\Z}{\IntReprI} \multiplication \X$
		\Statex
		\Return $\partderiv{\Z}{\W}$ 
	\end{algorithmic} 
\end{algorithm}

\section{Theoretical Analysis of Periodic Sparse Filtering for Covariate Shift Adaptation \label{sec:Theory-PSF}}

\subsection{Proof of Theorem 2 \label{app:Proof-of-Theorem-2}}

\setcounter{theorem}{1}
\begin{theorem}
	Let $\xone \in \origspace$ be a point in the original space and let $\zone \in \learnspace$ be its corresponding representation learned by PSF. Then there is an infinite set of points $\x \in \origspace$ that map onto the same representation $\zone$. The set of the points $\x \in \origspace$ built from $\xone$ with period $\W{}^{-1} \periodvector \pi$, where $\W$ is the weight matrix of PSF and $\periodvector$ is a vector of integer constants in $\domz$, is included in this set.
\end{theorem}

\textbf{Proof.}
The proof of this theorem is based on identifying the periodic filters defined by PSF in the original space and showing that points falling within these filters are mapped onto identical representations. The proof makes the following logical steps: (a) rigorous definition of the PSF computation; (b-e) back-computation through all the steps of PSF up to the input ($\lII$-normalization along the columns, $\lII$-normalization along the rows, non-linearity, linear projection).

(a) Let us consider two points in the original space $\origspace$:
\begin{eqnarray*}
	\xone & = & \transp{\vectorize{\begin{array}{cccc} \DATAMATRIXelem{x}{1}{1} & \DATAMATRIXelem{x}{1}{2} & \dots & \DATAMATRIXelem{x}{1}{\nXfeatures} \end{array}}} \\
	\xtwo & = & \transp{\vectorize{\begin{array}{cccc} \DATAMATRIXelem{x}{2}{1} & \DATAMATRIXelem{x}{2}{2} & \dots & \DATAMATRIXelem{x}{2}{\nXfeatures} \end{array}}},
\end{eqnarray*}
and their corresponding representations in the learned space $\learnspace$ defined by PSF:
\begin{eqnarray*}
	\zone & = & \transp{\vectorize{\begin{array}{cccc} \DATAMATRIXelem{z}{1}{1} & \DATAMATRIXelem{z}{1}{2} & \dots & \DATAMATRIXelem{z}{1}{\nZfeatures} \end{array}}} \\
	\ztwo & = & \transp{\vectorize{\begin{array}{cccc} \DATAMATRIXelem{z}{2}{1} & \DATAMATRIXelem{z}{2}{2} & \dots & \DATAMATRIXelem{z}{2}{\nZfeatures} \end{array}}}.
\end{eqnarray*}
Let us also consider a version of PSF implemented using a strictly positive element-wise sine function: 
$\APPLIEDFUNCTION{PSF}{\x} = \lIIcol{\lIIrow{1+\epsilon+\APPLIEDFUNCTION{\sin}{\W \xone}}}$.

Finally, let us assume that the two learned representations are identical, that is $\zone=\ztwo$. 

(b) By definition of PSF, $\zone=\ztwo$ implies:
\begin{eqnarray*}
	\lIIcol{\ftildeone} & = & \lIIcol{\ftildetwo} \\
	\frac{\ftildeonej}{\sqrt{\sumjfeatZ \ftildeonej^2}} & = &
	\frac{\ftildetwoj}{\sqrt{\sumjfeatZ \ftildetwoj^2}}, \\	
\end{eqnarray*}
where $\ftilde = \transp{\vectorize{\begin{array}{cccc} \DATAMATRIXelem{\tilde{f}}{i}{1} & \DATAMATRIXelem{\tilde{f}}{i}{2} & \dots & \DATAMATRIXelem{\tilde{f}}{i}{\nZfeatures} \end{array}}}$ is the intermediate output of PSF which is defined as $\Ftilde = \lIIrow{1+\epsilon+\APPLIEDFUNCTION{\sin}{\W \xone}}$.
Now, for the $\lII$-normalizations along the columns to be equal, it must hold that:
\[
\transp{\vectorize{\begin{array}{cccc}
		\frac{\DATAMATRIXelem{\tilde{f}}{1}{1}}{d_1} & 
		\frac{\DATAMATRIXelem{\tilde{f}}{1}{2}}{d_1} & \dots & \frac{\DATAMATRIXelem{\tilde{f}}{1}{\nZfeatures}}{d_1}\end{array}}} =
\transp{\vectorize{\begin{array}{cccc}
		\frac{\DATAMATRIXelem{\tilde{f}}{2}{1}}{d_2} &
		\frac{\DATAMATRIXelem{\tilde{f}}{2}{2}}{d_2} & \dots & \frac{\DATAMATRIXelem{\tilde{f}}{2}{\nZfeatures}}{d_2}\end{array}}},
\]
where $d_i=\sqrt{\sumjfeatZ\ftildeij^2}$ is a sample-dependent scaling factor. Therefore, it follows that $\zone=\ztwo$ if and only if $\ftildeone=\lambda\ftildetwo$, for $\lambda \in \domR$.

%However, because of the normalization imposed by $\lII$-normalizations along the rows we have that:
%\[
%\DATAMATRIXelem{\tilde{f}}{1}{j}^2 + \DATAMATRIXelem{\tilde{f}}{2}{j}^2 = 1 - \alpha_j, \,\,\,\, \forall 1\leq j \leq \nZfeatures, 
%\]
%where $\alpha_j = \sum_{i=3}^{\nXsamples} \DATAMATRIXelem{\tilde{f}}{i}{j}^2$ is a positive constant accounting for the contribution of all the other samples to the normalization along the feature $j$.
%Because of the identity $\ftildeone=\lambda\ftildetwo$, we can rewrite:
%\begin{eqnarray*}
%	\lambda^2 \DATAMATRIXelem{\tilde{f}}{2}{j}^2 + \DATAMATRIXelem{\tilde{f}}{2}{j}^2 & = & 1 - \alpha_j \\
%	\DATAMATRIXelem{\tilde{f}}{2}{j}^2 (1 + \lambda^2) & = & 1 - \alpha_j, \\	
%\end{eqnarray*}

(c) By definition of PSF, $\ftildeone=\lambda\ftildetwo$ implies:
\begin{eqnarray*}
	\lIIrow{\fone} & = & \lambda \lIIrow{\ftwo} \\
	\frac{\fonej}{\sqrt{\sumisamples \fij^2}} & = & \lambda
	\frac{\ftwoj}{\sqrt{\sumisamples \fij^2}},
\end{eqnarray*}
where $\f = \transp{\vectorize{\begin{array}{cccc} \DATAMATRIXelem{f}{i}{1} & \DATAMATRIXelem{f}{i}{2} & \dots & \DATAMATRIXelem{f}{i}{\nZfeatures} \end{array}}}$ is the intermediate output of PSF defined as $\F=1+\epsilon+\APPLIEDFUNCTION{\sin}{\W \xone}$.
Now, for the $\lII$-normalizations along the rows to be equal, it must hold that:
\[
\transp{\vectorize{\begin{array}{cccc}
		\frac{\DATAMATRIXelem{f}{1}{1}}{t_1} & 
		\frac{\DATAMATRIXelem{f}{1}{2}}{t_2} & \dots & \frac{\DATAMATRIXelem{f}{1}{\nZfeatures}}{t_\nZfeatures}\end{array}}} = \lambda
\transp{\vectorize{\begin{array}{cccc}
		\frac{\DATAMATRIXelem{f}{2}{1}}{t_1} &
		\frac{\DATAMATRIXelem{f}{2}{2}}{t_2} & \dots & \frac{\DATAMATRIXelem{f}{2}{\nZfeatures}}{t_\nZfeatures}\end{array}}},
\]
where ${t_j}=\sqrt{\sumisamples\fij^2}$ is a feature-dependent scaling factor. Therefore, it follows that $\ftildeone=\lambda\ftildetwo$ if and only if $\fone= \lambda\ftwo$. 

(d) By definition of PSF, $\fone= \lambda \ftwo$ implies:
\[
1+\epsilon+\APPLIEDFUNCTION{\sin}{\DATAMATRIXrow{h}{1}{}} = \APPLIEDFUNCTION{\lambda}{1+\epsilon+\APPLIEDFUNCTION{\sin}{\DATAMATRIXrow{h}{2}{}}},
\]
where $\DATAMATRIXrow{h}{i}{} = \transp{\vectorize{\begin{array}{cccc} \DATAMATRIXelem{h}{i}{1} & \DATAMATRIXelem{h}{i}{2} & \dots & \DATAMATRIXelem{h}{i}{\nZfeatures} \end{array}}}$ is the intermediate output of PSF defined as $\DATAMATRIX{H}{}=\W\X$.
Now, in order to prove our statement about the set of points $\x \in \origspace$ built from $\xone$ with period $\W{}^{-1} \periodvector \pi$, let us consider the case where $\lambda=1$:
\begin{eqnarray*}
	1+\epsilon+\APPLIEDFUNCTION{\sin}{\DATAMATRIXrow{h}{1}{}} & = & 1+\epsilon+\APPLIEDFUNCTION{\sin}{\DATAMATRIXrow{h}{2}{}}\\
	\APPLIEDFUNCTION{\sin}{\DATAMATRIXrow{h}{1}{}} & = & \APPLIEDFUNCTION{\sin}{\DATAMATRIXrow{h}{2}{}}.
\end{eqnarray*}
For the applications of the sinusoidal function to be equal, it must hold that:
\begin{eqnarray*}
	\APPLIEDFUNCTION{\sin}{\DATAMATRIXrow{h}{1}{}} & = & \APPLIEDFUNCTION{\sin}{\DATAMATRIXrow{h}{2}{}}\\
	\DATAMATRIXrow{h}{1}{} & = & \APPLIEDFUNCTION{\arcsin}{\APPLIEDFUNCTION{\sin}{\DATAMATRIXrow{h}{2}{}}}\\
	\DATAMATRIXrow{h}{1}{} & = & \DATAMATRIXrow{h}{2}{} + \periodvector\pi,
\end{eqnarray*}
where $\periodvector$ is a vector of feature-dependent periodic factors in $\domz$. 

(e) By definition of PSF, $\DATAMATRIXrow{h}{1}{} = \DATAMATRIXrow{h}{2}{} + \periodvector\pi$ implies:
\begin{eqnarray*}
	\W\xone & = & \W\xtwo + \periodvector\pi\\
	\xone & = & \xtwo + \W{}^{-1}\periodvector\pi.
\end{eqnarray*}
Thus, there are infinite points $\x \in \origspace$ built from $\xone$ with period $\W{}^{-1} \periodvector\pi$ that maps onto the same representation $\zone$. $\QED$

\section{Experimental Validation \label{app:Experiments}}

The six CSA classification systems used in our experiments were implemented and configured as follows:

\begin{enumerate}[label=(\roman*)]
	
	\item  \emph{SVM (without CSA)}: SVM is implemented using the \emph{scikit} implementation\footnote{\url{http://scikit-learn.org/}}.
	
	\item \emph{SF+SVM}: SF is implemented using the code provided by Ngiam\footnote{\url{https://github.com/jngiam/sparseFiltering}} \citep{Ngiam2011}. SVM is implemented as for system (i).
	
	\item \emph{PSF+SVM}: PSF is implemented using our own code\footnote{\url{https://github.com/FMZennaro/PSF}}.
	SVM is implemented as for system (i).  
	
	\item \emph{IW+LSPC}: the integrated IW+LSPC is implemented using the code provided by Hachiya\footnote{\url{http://www.ms.k.u-tokyo.ac.jp/software.html\#IWLSPC}} \citep{Hachiya2012}.
	
	\item \emph{SSA+SVM}: SSA is implemented using the code provided by Fernando\footnote{\url{http://users.cecs.anu.edu.au/~basura/DA_SA/}} \citep{Fernando2013}. SVM is implemented using the \emph{Matlab} implementation.
	
	\item \emph{DAE+SVM}: DAE is implemented using the code provided by Mitra\footnote{\url{https://github.com/rajarsheem/libsdae}}.
	SVM is implemented as for system (i).
	
\end{enumerate}

\subsection{Synthetic Data Experiments \label{app:Synthetic}}

\subsubsection*{Data Generation}

The four synthetic datasets were generated as follows:\\

(a) \emph{Radial dataset}: 
$\mathbf{X^{tr}}$ consists of 500 samples from $p\left(X^{tr}\right)\sim\mathcal{N}\left(\left[\begin{array}{c}
0.5\\
0
\end{array}\right];\left[\begin{array}{cc}
0.2 & 0\\
0 & 0.5
\end{array}\right]\right)$;
$\mathbf{X^{tst}}$ consists of 500 samples from $p\left(X^{tst}\right)\sim\mathcal{N}\left(\left[\begin{array}{c}
-0.5\\
0
\end{array}\right];\left[\begin{array}{cc}
0.2 & 0\\
0 & 0.5
\end{array}\right]\right)$;
$\mathbf{X^{tar}}$ consists of 250 samples from $p\left(X^{tst}\right)$;
$p\left(Y \vert X \right)$ is described by the following deterministic function 
$f\left(\mathbf{x}_i\right)\begin{cases}
1 & \textrm{if}\,\,\,\left|x_{i,1}\right|>\left|x_{i,2}\right|\\
0 & \textrm{otherwise}
\end{cases}$,
which defines two cones centered on the \emph{x}-axis.\\

(b) \emph{Periodic dataset}: 
$\mathbf{X^{tr}}$ consists of 500 samples from $p\left(X^{tr}\right)\sim\mathcal{N}\left(\left[\begin{array}{c}
2\pi\\
0
\end{array}\right];\left[\begin{array}{cc}
2 & 0\\
0 & 0.5
\end{array}\right]\right)$;
$\mathbf{X^{tst}}$ consists of 500 samples from $p\left(X^{tst}\right)\sim\mathcal{N}\left(\left[\begin{array}{c}
-2\pi\\
0
\end{array}\right];\left[\begin{array}{cc}
2 & 0\\
0 & 0.5
\end{array}\right]\right)$;
$\mathbf{X^{tar}}$ consists of 250 samples from $p\left(X^{tst}\right)$;
$p\left(Y \vert X \right)$ is described by the following deterministic function 
$f\left(\mathbf{x}_i\right)\begin{cases}
1 & \textrm{if}\,\,\,\sin\left|x_{i,1}\right|>0\\
0 & \textrm{otherwise}
\end{cases}$,
which defines a periodic pattern perpendicular to the \emph{x}-axis.
\\

(c) \emph{Smooth dataset}: 
$\mathbf{X^{tr}}$ consists of 250 samples from $p\left(X^{tr1}\right)\sim\mathcal{N}\left(\left[\begin{array}{c}
2\\
3
\end{array}\right];\left[\begin{array}{cc}
1 & 0\\
0 & 2
\end{array}\right]\right)$ and 250 samples from $p\left(X^{tr2}\right)\sim\mathcal{N}\left(\left[\begin{array}{c}
-2\\
3
\end{array}\right];\left[\begin{array}{cc}
1 & 0\\
0 & 2
\end{array}\right]\right)$;
$\mathbf{X^{tst}}$ consists of 250 samples from $p\left(X^{tst1}\right)\sim\mathcal{N}\left(\left[\begin{array}{c}
3\\
-1
\end{array}\right];\left[\begin{array}{cc}
1 & 0\\
0 & 1
\end{array}\right]\right)$ and 250 samples from $p\left(X^{tst2}\right)\sim\mathcal{N}\left(\left[\begin{array}{c}
0\\
-1
\end{array}\right];\left[\begin{array}{cc}
1 & 0\\
0 & 1
\end{array}\right]\right)$;
$\mathbf{X^{tar}}$ consists of 125 samples from $p\left(X^{tst1}\right)$ and 125 samples from $p\left(X^{tst2}\right)$;
$p\left(Y \vert X \right)$ is described by the deterministic function
$f\left(\mathbf{x}_i\right)\begin{cases}
1 & \textrm{if}\,\,\,\frac{1+\tanh\left(x_{i,1}+\min\left(0,x_{i,2}\right)\right)}{2}>0.5\\
0 & \textrm{otherwise}
\end{cases}$. This dataset is the same as the one studied in \citet{Hachiya2012}.\\

(d) \emph{Diagonal dataset}: 
$\mathbf{X^{tr}}$ consists of 500 noisy samples taken along the diagonal of the first quadrant, $x_{i,1}^{\mathbf{tr}}\sim Unif\left(\left[0,3\right]\right)$ and $x_{i,2}^{\mathbf{tr}}=x_{i,1}^{\mathbf{tr}}+\mathcal{N}\left(0,0.2\right)$;
$\mathbf{X^{tst}}$ consists of 500 noisy samples taken along the diagonal of the fourth quadrant, $x_{i,1}^{\mathbf{tst}}\sim Unif\left(\left[-3,0\right]\right)$ and $x_{i,2}^{\mathbf{tst}}=x_{i,1}^{\mathbf{tst}}+\mathcal{N}\left(0,0.2\right)$;
$\mathbf{X^{tar}}$ consists of 250  noisy samples taken along the diagonal of the fourth quadrant as $\mathbf{X^{tst}}$;
$p\left(Y \vert X \right)$ is described by the deterministic function 
$f\left(\mathbf{x}_i\right)\begin{cases}
1 & \textrm{if}\,\,\,\left|x_{i,2}\right|>1.5\\
0 & \textrm{otherwise}
\end{cases}$.

\subsubsection*{Experimental Setup \label{settings1}}

The six CSA classification systems are configured as follows:

\begin{enumerate}[label=(\roman*)]
	
	\item \emph{SVM (with no CSA)}: a linear SVM is trained on $\left\{ \mathbf{X^{tr}},\mathbf{Y^{tr}}\right\}$, using a fixed penalty $C=1$.
	
	\item \emph{SF+SVM}: SF is trained on $\left\{ \mathbf{X^{tr}},\mathbf{X^{tar}}\right\}$ for 500 iterations; we set the learned dimensionality to 2.
	SVM is trained as for system (i).
	
	\item \emph{PSF+SVM}: PSF is trained on $\left\{ \mathbf{X^{tr}},\mathbf{X^{tar}},\mathbf{Y^{tr}}\right\}$ for 500 iterations; we set the learned dimensionality to 2 (divided evenly between the two classes), the non-linearity to sine, and $\lambda=1.0$.
	SVM is trained as for system (i).  
	
	\item \emph{IW+LSPC}: IW is trained on $\left\{ \mathbf{X^{tr}},\mathbf{X^{tar}}\right\}$ using the pre-set \emph{uLSIF} algorithm with 250 basis, candidate $\sigma=\{0.1,0.2,0.5,1,2,3\}$ and candidate $\lambda=\{0.001, 0.003, 0.01, 0.03, 0.1,$ $0.3, 1, 3, 10\} $. LSPC is trained on $\left\{ \mathbf{X^{tr}},\mathbf{Y^{tr}}\right\}$ with the same pre-set candidate $\sigma=\{0.1,0.2,0.5,1,2,3\}$ and candidate $\lambda=\{0.1, 0.17, 0.32, 0.56, 1\}.$
	
	\item \emph{SSA+SVM}: SSA is trained on $\left\{ \mathbf{X^{tr}},\mathbf{X^{tar}}\right\}$ using 2 PCA components. SVM is trained on $\left\{ \mathbf{X^{tr}},\mathbf{Y^{tr}}\right\}$, using a penalty $C=3$ to achieve the same baseline results as systems (i)-(iii).
	
	\item \emph{DAE+SVM}: DAE is trained on $\left\{ \mathbf{X^{tr}},\mathbf{Y^{tr}}\right\}$ for $10000$ epochs with mini-batches of size $50$ and learning rate of $0.001$; we set the learned dimensionality to $1$, the non-linearity to sigmoid, and the noise to Gaussian $\mathcal{N}\left(0,0.1\right)$.
	SVM is trained as for system (i).
	
\end{enumerate}

\subsection{Real-World Data Experiments \label{app:Real-World}}

\subsubsection*{Experimental Setup}

All the recordings are pre-processed into standard feature representations using the open-source platform OpenSMILE\footnote{\url{http://audeering.com/technology/opensmile/}}. For each 1-second sample we compute features on 2 \emph{domains} (raw, delta), extracting 12 \emph{descriptors} (12 Mel-frequency cepstrum coefficient) and computing 3 \emph{statistical operators} (mean, standard deviation and range), for a total of 72 features. Labels are aligned along the valence dimension using a standard mapping in the ESR community \citep{Schuller2010} as specified in Table \ref{tab:ESR-mapping}. 

All the samples are normalized per speaker, as in \citet{Schuller2010}. Training and test data are upsampled using a simple re-sampling with repetition procedure, as in \citet{Zhang2013}.\\

\begin{table*}
	\caption{Mapping of labels specific to each data set onto binary valence classes. \label{tab:ESR-mapping}}
	\begin{centering}
		\begin{tabular}{ccc}
			\hline
			{} & 
			\textbf{\emph{\footnotesize{Positive Valence}}} &
			\textbf{\emph{\footnotesize{Negative Valence}}} 
			\tabularnewline 
			\hline
			\textbf{\emph{\footnotesize{EMODB}}} & joy, neutral & anger, boredom, disgust, fear, sadness \tabularnewline
			\textbf{\emph{\footnotesize{DES}}}  & happiness, neutral, surprise & angry, sadness \tabularnewline
			\textbf{\emph{\footnotesize{VAM}}}  & $valence>0$ & $valence<0$  \tabularnewline
			\textbf{\emph{\footnotesize{eNT}}} & joy, surprise & anger, disgust, fear, sadness  \tabularnewline
			\hline
			
		\end{tabular}
		\par\end{centering}
	
\end{table*}	

\begin{table*}
	\caption{Best hyper-parameter configurations chosen by model selection. \label{tab:ModelSelection}}
	\begin{centering}
		
		\begin{tabular}{>{\centering}p{1.0cm}>{\centering}p{2cm}>{\centering}p{2.2cm}>{\centering}p{2.8cm}}
			
			\hline 
			& \textbf{\emph{\footnotesize{}noCSA+SVM}} & \textbf{\emph{\footnotesize{}SF+SVM}} & 
			\textbf{\emph{\footnotesize{}PSF+SVM}}  
			\tabularnewline
			
			\hline 
			\textbf{\emph{\footnotesize{}EMODB}} & 
			{\footnotesize{}SVM: $C=7.5\cdot10^{-5}$} & 
			
			{\footnotesize{}SF: $L=50$}{\footnotesize \par}
			{\footnotesize{}SVM: $C=0.075$} & 
			
			{\footnotesize{}PSF: $g=\cos\left(\right)$}{\footnotesize \par}
			{\footnotesize{}PSF: $L=80$}{\footnotesize \par}
			{\footnotesize{}PSF: $\lambda=1.8$}{\footnotesize \par}
			{\footnotesize{}SVM: $C=7.5\cdot10^{-5}$} 
			\tabularnewline
			
			\hline 
			\textbf{\emph{\footnotesize{}DES}} & {\footnotesize{}SVM: $C=5\cdot10^{-5}$} &
			
			{\footnotesize{}SF: $L=120$}{\footnotesize \par}
			{\footnotesize{}SVM: $C=0.001$} & 
			
			{\footnotesize{}PSF: $g=\sin\left(\right)$}{\footnotesize \par}
			{\footnotesize{}PSF: $L=100$}{\footnotesize \par}
			{\footnotesize{}PSF: $\lambda=1.6$}{\footnotesize \par}
			{\footnotesize{}SVM: $C=1\cdot10^{-4}$}
			\tabularnewline
			
			\hline 
			\textbf{\emph{\footnotesize{}eNT}} & 
			
			{\footnotesize{}SVM: $C=0.01$} & 
			
			{\footnotesize{}SF: $L=50$}{\footnotesize \par}
			{\footnotesize{}SVM: $C=1.0$} & 
			
			{\footnotesize{}PSF: $g=\sin\left(\right)$}{\footnotesize \par}
			{\footnotesize{}PSF: $L=120$}{\footnotesize \par}
			{\footnotesize{}PSF: $\lambda=1.8$}{\footnotesize \par}
			{\footnotesize{}SVM: $C=7.5\cdot10^{-5}$}
			\tabularnewline

			\hline 
			& \textbf{\emph{\footnotesize{}IW+LSPC}} &
			\textbf{\emph{\footnotesize{}SSA+SVC}} &
			\textbf{\emph{\footnotesize{}DAE+SVM}} 
			\tabularnewline
			
			\hline 
			\textbf{\emph{\footnotesize{}EMODB}} & 
							
			{\footnotesize{}LSPC: $\sigma=3$}{\footnotesize \par}
			{\footnotesize{}LSPC: $\lambda=0.1$}{\footnotesize \par} &
			
			{\footnotesize{}SSA: $P=50$}{\footnotesize \par}
			{\footnotesize{}SVM:  $C=7.5\cdot10^{-5}$}{\footnotesize \par} & 
			
			{\footnotesize{}DAE: $f,g=\textnormal{sigm}()$}{\footnotesize \par}
			{\footnotesize{}DAE: $L=70$}{\footnotesize \par}
			{\footnotesize{}DAE: $\sigma=0.1$}{\footnotesize \par}
			{\footnotesize{}DAE: $\eta=0.005$}
			{\footnotesize{}SVM: $C=0.00025$}
			\tabularnewline
			
			\hline 
			\textbf{\emph{\footnotesize{}DES}} &
			
			{\footnotesize{}LSPC: $\sigma=3$}{\footnotesize \par}
			{\footnotesize{}LSPC: $\lambda=1$}{\footnotesize \par} &
			
			{\footnotesize{}SSA: $P=50$}{\footnotesize \par}
			{\footnotesize{}SVM:  $C=2.5\cdot10^{-4}$}{\footnotesize \par} &
			
			{\footnotesize{}DAE: $f,g=\textnormal{sigm}()$}{\footnotesize \par}
			{\footnotesize{}DAE: $L=100$}{\footnotesize \par}
			{\footnotesize{}DAE: $\sigma=0.01$}{\footnotesize \par}
			{\footnotesize{}DAE: $\eta=0.005$}
			{\footnotesize{}SVM: $C=0.0001$}				
			\tabularnewline
			\hline 
			\textbf{\emph{\footnotesize{}eNT}} &
			
			{\footnotesize{}LSPC: $\sigma=0.5$}{\footnotesize \par}
			{\footnotesize{}LSPC: $\lambda=0.3$}{\footnotesize \par} &
			
			{\footnotesize{}SSA: $P=50$}{\footnotesize \par}
			{\footnotesize{}SVM:  $C=7.5\cdot10^{-5}$}{\footnotesize \par} &
			
			{\footnotesize{}DAE: $f,g=\textnormal{sigm}()$}{\footnotesize \par}
			{\footnotesize{}DAE: $L=120$}{\footnotesize \par}
			{\footnotesize{}DAE: $\sigma=0.01$}{\footnotesize \par}
			{\footnotesize{}DAE: $\eta=0.001$}
			{\footnotesize{}SVM: $C=0.005$}				
			\tabularnewline
			\hline

		\end{tabular}
		\par\end{centering}
\end{table*}

In order to perform model selection over a reasonable set of values for the hyper-parameters of each algorithm, the six CSA classification systems are configured as follows:

\begin{enumerate}[label=(\roman*)]
	
	\item \emph{SVM}: a linear SVM is trained on $\left\{ \mathbf{X^{tr}},\mathbf{Y^{tr}}\right\}$. Model selection is performed on a single hyper-parameter: the soft cost $C$ is chosen in the set $\{2.5\cdot10^{-5}, 5\cdot10^{-5}, 7.5\cdot10^{-5}, ..., 0.5, 0.75, 1.0\}$ following \citet{Eyben2016}.
	
	\item \emph{SF+SVM}: SF is trained on $\left\{ \mathbf{X^{tr}},\mathbf{X^{tar}}\right\}$ using an early stopping criterion (see below). Model selection is performed on a single hyper-parameter: the learned features $L$ is chosen in the set $\{20,50,80,100,120\}$ in order to explore a coarse-grained grid space around the original number of features $M=72$.
	SVM is trained as for system (i).
	
	\item \emph{PSF+SVM}: PSF is trained on $\left\{ \mathbf{X^{tr}},\mathbf{X^{tar}},\mathbf{Y^{tr}}\right\}$ using an early stopping criterion (see below).
	Model selection is performed on three hyper-parameters: the non-linearity $g$ chosen in the set $\{\sin(), \cos()\}$, the learned feature $L$ in the set $\{20,50,80,100,120\}$ as before, the loss parameter $\lambda$ in the set $\{0.8, 0.9, 0.95, 1.0, 1.05, 1.1, 1.15, 1.2, 1.3,1.4, 1.6, 1.8,$ $2.0, 2.2, 2.4, 2.6, 2.8\}$ in order to explore a finer-grained grid space around the value $\lambda=1$. The dimensionality of the learned space is given by the number of features $L$, evenly divided between the two classes, plus a fixed number of $10$ features to account for unlabelled samples.
	SVM is trained as for system (i).  
	
	\item \emph{IW+LSPC}: IW is trained on $\left\{ \mathbf{X^{tr}},\mathbf{X^{tar}}\right\}$ setting the number of basis to the cardinality of the target set $\mathbf{X^{tar}}$. Model selection is performed on two hyper-parameters: the candidate $\sigma$ chosen in the set $\{0.1,0.2,0.5,1,2,3\}$ and the candidate $\lambda$ in the set $\lbrace 0.001, 0.003, 0.01, 0.03, 0.1, 0.3, 1, 3, 10 \rbrace$ following \citet{Hachiya2012}.
	LSPC is trained on $\left\{ \mathbf{X^{tr}},\mathbf{Y^{tr}}\right\}$. Model selection is performed on two hyper-parameter: the candidate $\sigma$ chosen in the set $\{0.1,0.2,0.5,1,2,3\}$ and the candidate $\lambda$ in the set $\{0.1, 0.17, 0.32, 0.56, 1\}$ following \citet{Hachiya2012}.
	
	\item \emph{SSA+SVM}: SSA is trained on $\left\{ \mathbf{X^{tr}},\mathbf{X^{tar}}\right\}$.
	Model selection is performed on a single hyper-parameter: the number of PCA components $P$ chosen in the set $\{35, 50, 70\}$ in order to explore a coarse-grained grid below the original number of features $M=72$.
	SVM is trained as for system (i).	
	
	\item \emph{DAE+SVM}: DAE is trained on $\left\{ \mathbf{X^{tr}},\mathbf{Y^{tr}}\right\}$ for $10000$ epochs and the noise set to a Gaussian with zero mean. Model selection is performed on four hyper-parameters: the non-linearity (for both encoding and decoding) is chosen in the set $\{\textnormal{sigmoid}(), \textnormal{tanh}()\}$ following the standard in the neural network literature, the learned features $L$ in the set $\{70,100,120\}$ in order to explore a coarse-grained grid equal or above the original number of features $M=72$, the variance of the noise $\sigma$ in the set $\{0.01,0.05,0.1\}$ and the learning rate $\eta$ in the set $\{0.0005,0.001,0.005\}$ following the standard in the neural network literature.
	
\end{enumerate}

\subsubsection*{Experimental Results}

Table \ref{tab:ModelSelection} reports the hyper-parameter configurations selected by model selection.

\bibliographystyle{spbasic}      % basic style, author-year citations
\bibliography{references}   % name your BibTeX data base

\end{document}